\documentclass[11pt]{article}
\usepackage{jheppub}
\usepackage{amssymb,amsfonts}
\usepackage{mathrsfs}
\usepackage{subcaption}
\usepackage{graphicx}
\usepackage{booktabs}
\let\savenumberline\numberline
\def\numberline#1{\savenumberline{#1.}}
\makeatletter
\renewcommand{\@seccntformat}[1]{\csname the#1\endcsname.\,\,}
\makeatother
\newcommand\secref[1]{{\S\ref{#1}}}
\newcommand\appref[1]{{Appendix~\ref{#1}}}
\newcommand\figref[1]{{Figure~\ref{#1}}}
\newcommand\tabref[1]{{Table~\ref{#1}}}

\newcommand{\CN}{{\cal N}}

\renewcommand{\tilde}[1]{\widetilde{#1}}
\renewcommand{\hat}[1]{\widehat{#1}}
\newcommand{\be}{\begin{equation}}
\newcommand{\ee}{\end{equation}}
\newcommand{\bea}{\begin{eqnarray}}
\newcommand{\eea}{\end{eqnarray}}

\newcommand{\R}{\mathbb{R}}

\newcommand{\norm}[1]{\left\lVert #1 \right\rVert}

\newcommand{\Attn}{\mathrm{Attn}}
\newcommand{\MLP}{\mathrm{MLP}}
\newcommand{\SNR}{\mathrm{SNR}}
\newcommand{\spec}{\mathrm{spec}}

\makeatletter
\def\@fpheader{\relax}
\makeatother
\usepackage{graphicx}
\usepackage{latexsym}
\title{Interpreting the Synchronization Gap: The Hidden Mechanism Inside Diffusion Transformers }
\author[a,b,1]{Emil Albrychiewicz,}
\author[a,b,c,1]{Andr\'{e}s Franco Valiente,}
\author[d]{Li-Ching Chen,}
\author[e]{Viola Zixin Zhao}
\affiliation[a]{Leinweber Institute for Theoretical Physics and Department of Physics,\\
University of California, Berkeley, CA, 94720-7300, USA}

\affiliation[b]{Theoretical Physics Group, Lawrence Berkeley National Laboratory,\\
Berkeley, CA 94720-8162, USA}

\affiliation[c]{Department of Radiation Oncology, University of California, San Francisco}

\affiliation[d]{
Computational Precision Health, University of California, San Francisco
 }
\affiliation[e]{Department of Physics,\\
University of California, Berkeley, CA 94720, USA}

\affiliation[1]{These authors contributed equally.}

\emailAdd{ealbrych@berkeley.edu}
\emailAdd{andresfranco@berkeley.edu}
\emailAdd{liching\_chen@berkeley.edu}
\emailAdd{zhaozixin@berkeley.edu}

\abstract{
Recent theoretical models of diffusion processes, conceptualized as coupled Ornstein-Uhlenbeck systems, predict a hierarchy of interaction timescales, and consequently, the existence of a synchronization gap between modes that commit at different stages of the reverse process. However, because these predictions rely on continuous time and analytically tractable score functions, it remains unclear how this phenomenology manifests in the deep, discrete architectures deployed in practice. In this work, we investigate how the synchronization gap is mechanistically realized within pretrained Diffusion Transformers (DiTs). We construct an explicit architectural realization of replica coupling by embedding two generative trajectories into a joint token sequence, modulated by a symmetric cross attention gate with variable coupling strength $g$. Through a linearized analysis of the attention difference, we show that the replica interaction decomposes mechanistically. We empirically validate our theoretical framework on a pretrained DiT-XL/2 model by tracking commitment and per layer internal mode energies. Our results reveal that: (1) the synchronization gap is an intrinsic architectural property of DiTs that persists even when external coupling is turned off; (2) as predicted by our spatial routing bounds, the gap completely collapses under strong coupling $g\rightarrow1$; (3) the gap is strictly depth localized, emerging sharply only within the final layers of the Transformer; and (4) global, low frequency structures consistently commit before local, high frequency details. Ultimately, our findings provide a mechanistic interpretation of how Diffusion Transformers resolve generative ambiguity, isolating speciation transitions to the terminal layers of the network. 
}

\begin{document}

\maketitle

\section{Introduction}

Diffusion models generate data by learning to reverse a stochastic noising process, progressively transforming Gaussian noise into structured samples \cite{sohl2015deep, ho2020denoising, song2020score, lai2025principles}. Among current architectures, Diffusion Transformers (DiTs) \cite{peebles2023scalable} have emerged as the foundational standard for generative modeling. By replacing the rigidly structured convolutional U-Net \cite{ronneberger2015u} with a highly scalable sequence of Transformer blocks \cite{vaswani2017attention} that operate on modality agnostic patchified latent tokens \cite{rombach2022high}, DiTs have rapidly expanded beyond their origins in high fidelity image synthesis. This architecture now powers SOTA multimodal systems across highly specialized domains ranging from healthcare applications \cite{kazerouni2023diffusion}, spatiotemporal video generation \cite{ho2022imagenvideo, brooks2024video}, and 3D molecular drug design \cite{alakhdar2024diffusion} to complex statistical applications in causal inference \cite{sanchez2022diffusion, ma2024diffpo}. However, despite this unprecedented empirical success, the internal mechanisms by which these models resolve generative ambiguity when transitioning from unstructured noise to specific, coherent representations remains poorly understood from an interpretability perspective. 

The necessity for interpretability \cite{linardatos2020explainable} in deep learning stems from a fundamental trade off between predictive capacity and human comprehension. In basic statistical models like linear regression, the mapping is intrinsically interpretable: every learned weight directly quantifies the influence of a feature on the output, allowing practitioners to easily audit decisions and extract true causal relationships. However, as architectures scale into deep, nonlinear neural networks, this transparency degrades into a black box. Overcoming this opacity is not merely a post hoc diagnostic exercise, but a critical engineering requirement where interpreting internal representations allows developers to debug failure modes, predict edge case behaviors, and design more efficient, reliable architectures. Furthermore, interpretability is a strict prerequisite for deployment in high stakes domains such as healthcare, where algorithmic accountability, bias mitigation, and safety are legally and ethically mandated \cite{goktas2025shaping, ennab2024enhancing}. This need is particularly acute in the natural sciences. As physicists, biologists, and chemists increasingly adopt foundational deep learning models, they require more than just a highly accurate predictive black box, they require mechanistic insight. 

A recent line of work based on nonequilibrium statistical physics \cite{raya2023spontaneous, biroli2024, kamb2024analytic, sclocchi2025phase} has begun to address these interpretability questions by identifying sharp dynamical transitions in the reverse generative process. In \cite{biroli2024}, the authors established that diffusion models trained on structured data distributions undergo two macroscopic phase transitions, a speciation time, at which the trajectory commits to a particular data mode, and a collapse time, at which it locks onto a specific training example. These transitions have since been characterized through replica analysis of Gaussian mixtures \cite{biroli2024}, extended to general class structures via free entropy criteria \cite{achilli2026theory}, and connected to entropic signatures observable in trained models \cite{handke2026entropic}. In the setting of coupled multimodal generation, \cite{albrychiewicz2026dynamical} showed that modeling the interaction between two coupled diffusion trajectories as a pair of Ornstein--Uhlenbeck (OU) processes \cite{uhlenbeck1930theory} reveals a synchronization gap. In the case of symmetric coupling, this is a temporal window during which the common eigenmode has speciated while the difference eigenmode has not. This gap arises from a spectral hierarchy in the interaction timescales and has been shown to depend on the coupling strength $g$. The coupled OU processes approach has also been recently extended by \cite{lu2026steering} for breaking detailed balance which can be used to accelerate generative process. 

These theoretical results, however, are formulated entirely in the language of continuous stochastic processes with analytically tractable score functions. In contrast, a pretrained Diffusion Transformer is not analytically tractable. It is a deep, discrete residual network in which the score function is implicitly defined by the composition of attention layers, pointwise nonlinearities, and adaptive normalization modules. The central question motivating this work is how the synchronization gap phenomenology is realized in the architecture of a Diffusion Transformer, and what mechanism is responsible for its existence? 

We answer this question both theoretically and empirically. On the theoretical side, we construct an explicit mapping from the coupled OU system into the self-attention mechanism of a pretrained DiT. By embedding two generation trajectories into a single token sequence and introducing a blockwise normalized, symmetric attention gate which depends on the coupling strength $g$, we obtain a controlled architectural realization of replica coupling \eqref{eqn:AttnBlocks}. Linearizing the resulting difference in attention output around the symmetric state, where two replicas are equivalent, we decompose it into two mechanistically distinct terms \eqref{eqn:AttnDelta}. The first term, which we refer to as a spatial routing, is the one where unperturbed attention kernel transports a perturbed value signal across token positions, and in the second, the perturbation enters through the softmax Jacobian of the attention weights themselves. We show that these two channels are suppressed by different functions of the coupling strength, $\tfrac{1-g}{1+g}$ and $\tfrac{1}{1+g}$, respectively, and argue that the first term is dominant for low frequency modes of replicas difference.

To move beyond the linear regime and to determine a speciation time, we model the local distribution of the replicas difference modes as a symmetric two component Gaussian mixture, following the approach of \cite{biroli2024, albrychiewicz2026dynamical} adapted to the discrete block structure. Projecting the resulting fixed point equation \eqref{eqn:fixed-point-vector} onto empirical eigenmodes of the initial difference covariance yields a scalar self consistency condition for each mode \eqref{eqn:scalar-self-consistency}, with a modewise speciation parameter that decomposes into an attention gated signal-to-noise ratio (SNR) \eqref{eqn:snr-def}. We also briefly mention how renormalization group flow \cite{wilson1975renormalization} approach can be used to analyze the Transformer architecture. The synchronization gap \eqref{eqn:snr-diff-scaling}, the difference in speciation times between leading and trailing modes is then shown to scale as $\mathcal{O}(\tfrac{1-g}{1+g})$ under an assumption that spatial routing term dominates, predicting its shrinkage at strong coupling. 

On the empirical side, we test these predictions on a pretrained DiT-XL/2 model \cite{peebles2023scalable} using two complementary experimental protocols. The first protocol \secref{sec:EmpProtI} measures when the model behaviorally commits by coupling two replicas for an initial steps and then letting them evolve independently, measuring the agreement of the final decoded images through feature space cosine similarity (using pretrained ResNet-50 encoder \cite{he2016deep}) and scale dependent pixel discrepancies. The second protocol \secref{sec:EmpProtII}, with a sweep across all 28 Transformer layers, identifies where this commitment is represented internally by tracking the empirical mode energies of the hidden state replica difference, evaluated at the speciation time identified by Protocol I.

\begin{figure}
    \centering
    \includegraphics[width=0.99\linewidth]{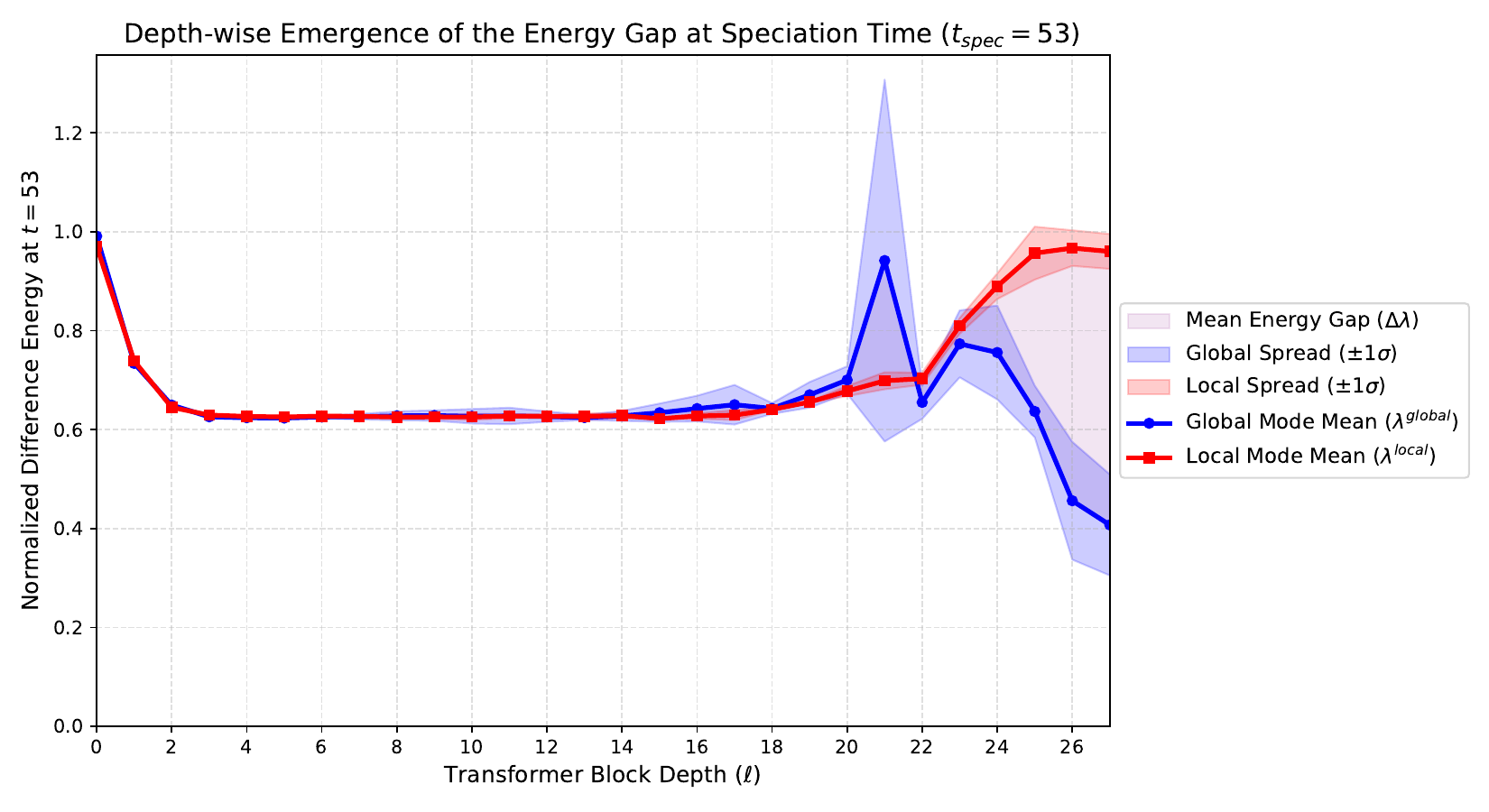}
    \caption{For coupling strength $g=0$ and speciation step $s=53$, we
    sweep across all Transformer layers  to evaluate the normalized fixed basis energies of leading and trailing internal difference
    modes \eqref{eqn:ProtIILead}--\eqref{eqn:ProtIITrail}. Even with coupling turned off, the gap is present at deep layers of DiT.}
    \label{fig:SweepG0}
\end{figure}

Our main empirical findings are:
\begin{enumerate}
    \item The synchronization gap exists in DiT even without coupling. At $g=0$, the Protocol II experiment reveals a clear separation between leading and trailing modes energies concentrated in the final layers of Transformer c.f., \figref{fig:SweepG0}. This demonstrates that the gap is an intrinsic property of pretrained DiT architecture, not just an artifact of the imposed coupling. 
    \item The gap collapses in the strong coupling region as predicted in theory section c.f., \figref{fig:SweepG01}--\figref{fig:SweepG09}. As $g$ increases from $0$ to $1$, the internal leading and trailing mode separation is progressively suppressed. This is consistent with theoretical prediction that, under the assumption of spatial routing term dominance, the spectral hierarchy collapses as $g\to 1$.
    \item The gap is depth localized. In the weakly coupled regime, the synchronization gap is near zero in early layers and exhibits a transient texture inversion in middle layers, where local features temporarily stabilize before global ones. The gap emerges sharply only within the last $\approx 5$ layers. This identifies the terminal layers as the site where the network performs
frequency based routing.
    \item Global structure commits before local detail. The scale dependent probe of Protocol I confirms that low frequency image structure stabilizes substantially earlier than high frequency details across all tested coupling strengths $g$ c.f., \figref{fig:ComitG0}--\figref{fig:ComitG09}.
\end{enumerate}

The remainder of this paper is organized as follows. In \secref{sec:Theory} we develop the theoretical framework, from the OU motivation through the attention gated propagator to the modewise SNR formula and gap collapse prediction. We follow with \secref{sec:Empirical} in which we describe the two empirical protocols and their implementation. In \secref{sec:Results} we present the experimental results. We conclude with \secref{sec:Conclusions} where we discuss the findings and their implications on recent training free acceleration methods and other applications.  In \appref{app:DerAttnDelta} we provide the detailed derivation of the attention difference between replicas, and in \appref{app:RoutingDom} we establish the spatial routing term dominance bound for low frequency modes. 

\section{Theoretical Framework}
\label{sec:Theory}

We develop an effective theoretical framework connecting the continuous statistical physics of coupled diffusion processes to the discrete  architecture of Diffusion Transformers. The goal is not to claim an exact equivalence between the OU description and a pretrained transformer, but rather to extract a testable mechanistic hypothesis. The synchronization gap phenomenology predicted in coupled diffusion theory is implemented in DiTs through spatially selective routing in self-attention, and can be revealed by controlled replica coupling. In contrast to the case discussed in \cite{albrychiewicz2026dynamical}, the synchronization gap that we consider is a temporal window between the commitment times of distinct projected modes of the replica difference channel. In the empirical analysis, these modes are defined as fixed principal directions of the initial empirical difference covariance, and the measured observables are their time dependent projected energies.
\subsection{Coupled Reverse Diffusion: Motivation and Limitations of the Linear Theory}
\label{sec:coupled-reverse}

Unless stated otherwise, we will consider the diffusion process as taking values in the latent space of dimension $d_z$. The variance preserving (VP) forward diffusion process is an Ornstein-Uhlenbeck (OU) process which has an Ito SDE realization as  
\begin{align}
    dz_t=-\frac{1}{2}\beta_tz_tdt+\sqrt{\beta_t}dW_t, 
\end{align}
where $z_0\sim p_{\text{data}}$, $\beta_t$ is the noise schedule and $W_t$ a standard Wiener 
process in $\R^{d_z}$. From a physics perspective, this SDE describes the overdamped Langevin dynamics of a particle coupled to a heat reservoir. Under this process, the conditional forward marginal given $z_0$ is
\begin{align}
    q(z_t \mid z_0)\sim\CN(\alpha_tz_0,\sigma_t^2I),
\end{align}
where 
\begin{align}
\alpha_t=e^{-\frac{1}{2}\int_0^t\beta_sds}, \quad \text{and} \quad \sigma_t^2=1-\alpha_t^2.
\end{align}

Following \cite{albrychiewicz2026dynamical}, we study two replicas $z_t^A, z_t^B\in \R^{d_z}$ which are two realizations/trajectories of an identical stochastic process. Here, the superscript $A,B$ are distinct labels, not a free index. These replicas are constructed in such a way that their reverse time dynamics are coupled with a symmetric relaxation matrix with a strength $g\geq 0$. In the continuous OU approximation, the coupled reverse process is
\begin{align}
\label{eqn:SymEqn1}
    dz_t^A&=\left[f(z_t^A,t)+g(z_t^A-z_t^B)\right]dt+\sqrt{\beta_t}  d\bar{W}_t^A,
    \\
    \label{eqn:SymEqn2}
    dz_t^B&=\left[f(z_t^B,t)+g(z_t^B-z_t^A)\right]dt+\sqrt{\beta_t} d\bar{W}_t^B,
\end{align}
where the reverse drift is
\begin{align}
    f(z_t,t)=-\frac{1}{2}\beta_tz_t-\beta_t\nabla_{z_t}\log p_t(z_t)
\end{align}
and $s(z_t,t)=\nabla_{z_t} \log p_t(z_t)$ is the score. The reverse process is initialized at the terminal time $t=T$ from the stationary Gaussian distribution $z_T^A, z_T^B \sim \mathcal{N}(0, I)$ with the convention that the time orientation flows backwards $dt<0$. We assume that the reverse Wiener processes $\bar{W}_t^A$ and $\bar{W}^B_t$ are independent white noise.

Similarly to the approach used in \cite{albrychiewicz2026dynamical}, we introduce common and difference modes
\begin{align}
    \label{eq:uv-coords}
    u_t = \frac{z_t^A + z_t^B}{\sqrt{2}}\,,\qquad
    v_t = \frac{z_t^A - z_t^B}{\sqrt{2}}\,.
\end{align}
However, in this case, we cannot use the exact score and these modes do not fully decouple the system given by equations \eqref{eqn:SymEqn1}-\eqref{eqn:SymEqn2}. Nevertheless, at the beginning of the reverse process, when noise level is high, we can approximate the marginal. If one approximates the marginal score by a single Gaussian $\CN(0,\Sigma_t)$ then the score is linear
\begin{align}
    \nabla_z \log p_t(z) \approx -\Sigma_t^{-1} z,
\end{align}
in that regime, the dynamics decouple
\begin{align}
    \mathrm{d}u_t &= \bigl[-\tfrac{1}{2}\beta_t\,u_t 
        + \beta_t\,\Sigma_t^{-1}\,u_t\bigr]\,\mathrm{d}t 
        + \sqrt{\tfrac{\beta_t}{2}}
        \,\mathrm{d}(\bar{W}_t^A + \bar{W}_t^B)\,,
    \label{eq:u-dynamics}\\[4pt]
    \mathrm{d}v_t &= \bigl[-\tfrac{1}{2}\beta_t\,v_t 
        + \beta_t\,\Sigma_t^{-1}\,v_t 
        + 2g\,v_t\bigr]\,\mathrm{d}t 
        + \sqrt{\tfrac{\beta_t}{2}}
        \,\mathrm{d}(\bar{W}_t^A - \bar{W}_t^B)\,.
    \label{eq:v-dynamics}
\end{align}
The common mode~$u_t$ evolves identically to a single uncoupled 
trajectory. The difference mode~$v_t$ acquires an additional 
restoring term $2g\,v_t$ that accelerates its drift toward zero mean hence acting as a restoring force that pulls the replicas together.

The linear theory predicts a hierarchy of damping rates across directions in latent space. However, by itself it does not identify those directions with spatial frequencies, nor can it generate branch formation. As long as the linearized operator remains stable, the unique fixed point for the difference mode is $v=0$ \footnote{For later discussion we suppress index $t$ for clarity.}. Hence the linear Gaussian theory is best understood as a motivation for a difference channel hierarchy, not yet as a complete account of speciation. 

This motivates two extensions. First, we want to implement an architectural realization of symmetric coupling inside DiT self-attention. Unlike a simple linear penalty $g(z^A - z^B)$, joint self-attention inherently introduces non-linearity through its softmax activation and multiplicative query-key interactions. By allowing the tokens of replica $A$ to attend to the tokens of replica $B$, the effective coupling strength becomes state dependent which brings us to the realm of nonlinear SDEs or SDEs with multiplicative noise. Second, from a distributional perspective, we want to consider the simplest non-linear marginal probability that breaks the linear score assumption and allows stable branch formation in the difference channel. To generate branch formation, the system must undergo a symmetry breaking bifurcation which would result in a nontrivial fixed point $v\neq 0$. This mathematically requires the drift to exhibit a localized instability at the origin, which can only occur if the score function $\nabla_z \log p(z)$ contains higher order nonlinearities.  

\subsection{Discrete DiT Model Architecture}
\label{sec:AttnGating}
To parameterize the reverse dynamics, we use the Diffusion Transformer (DiT) architecture \cite{peebles2023scalable}. A DiT operates on a sequence of latent tokens rather than on a spatial feature pyramid as in a convolutional U-Net. We begin with a noisy spatial latent
\begin{align}
z_t \in \mathbb{R}^{C \times H \times W},
\end{align}
where $C$, $H$, and $W$ denote the latent channel, height, and width dimensions. We partition $z_t$ into non-overlapping patches of size $p \times p$, assuming $p \mid H$ and $p \mid W$, flatten each patch, and linearly project it into $\mathbb{R}^{d_{\mathrm{model}}}$. This yields a token sequence of length $N$
\begin{align}
X_t \in \mathbb{R}^{N \times d_{\mathrm{model}}}, \qquad N = (H/p)(W/p).
\end{align}
The Transformer maps this sequence to an output sequence of the same length, conditioned on the diffusion timestep $t$ and external conditioning $c$ through adaptive layer normalization. A final linear decoder followed by an unpatchifying reshape maps the output tokens back to $\mathbb{R}^{C \times H \times W}$ to produce the noise prediction $\hat{\epsilon}_\theta(z_t,t,c)$. In the DiT-XL/2 configuration, $p=2$ and $d_{\mathrm{model}}=1152$.

Let $H_{0} = X_t$. Each Transformer block then takes the form
\begin{align}
\label{eqn:BlockUp1}
\tilde{H}_{\ell} &= H_{\ell} + \alpha_{\ell}\odot \Attn_{\ell}\!\bigl(\mathrm{adaLN}_{1,\ell}(H_{\ell}, t, c)\bigr), \\
\label{eqn:BlockUp2}
H_{\ell+1} &= \tilde{H}_{\ell} + \beta_{\ell} \odot \MLP_{\ell}\!\bigl(\mathrm{adaLN}_{2,\ell}(\tilde{H}_{\ell}, t, c)\bigr),
\end{align}
where $\Attn_{\ell}$ denotes multi-head self-attention, which mixes information across token positions, and $\MLP_{\ell}$ is a tokenwise feed-forward network acting independently on each token. The adaptive layer-normalization modules $\mathrm{adaLN}_{1,\ell}$ and $\mathrm{adaLN}_{2,\ell}$ shift and scale normalized activations using embeddings derived from the diffusion timestep $t$ and conditioning $c$. The gating vectors $\alpha_{\ell}, \beta_{\ell} \in \mathbb{R}^{d_{\mathrm{model}}}$ are likewise predicted from the conditioning embeddings, broadcast across token positions, and initialized at zero to stabilize training.

For a replica pair $(A,B)$, we concatenate the two token sequences along the token dimension,
\begin{align}
X_t = [X_t^A; X_t^B] \in \mathbb{R}^{2N \times d_{\mathrm{model}}}.
\end{align}
For each attention head, let $Q$, $K$, and $V$ denote the corresponding query, key, and value matrices, and let $d_h$ be the per-head hidden dimension. The pre-softmax attention logits are
\begin{align}
S = \frac{QK^\intercal}{\sqrt{d_h}}.
\end{align}
Under the replica concatenation, these decompose into block form as
\begin{align}
S = \begin{bmatrix}
S_{AA} & S_{AB} \\
S_{BA} & S_{BB}
\end{bmatrix},
\qquad
V =
\begin{bmatrix}
V_A \\
V_B
\end{bmatrix}.
\end{align}
Here $S_{AA}$ and $S_{BB}$ are the intra-replica attention logits, while $S_{AB}$ and $S_{BA}$ are the inter-replica logits. This decomposition will allow us to introduce attention gating by modulating the inter-replica interactions separately from the intra-replica ones.

A naive softmax over all $2N$ tokens would implicitly couple replicas through the partition function even when the coupling strength $g=0$, introducing a confound factor. To ensure a clean baseline, we compute attention weights with separate row wise softmax normalizations within each block
\begin{align}
\label{eqn:AttenMatrix}
    [A_{ij}]_{mn} = \frac{\exp([S_{ij}]_{mn})}{\sum_{k=1}^N \exp([S_{ij}]_{mk})},
\end{align}
where $m$ denotes query token and $n$ the key token within that block so $m,n\in\{1,2,\cdots,N\}$ and $i,j$ label $A,B$. These attention weights can be organized in terms of block matrices corresponding 

We realize the continuous coupling strength $g$ through a normalized mixture of intra replica and inter replica attention outputs
\begin{align}
\label{eqn:AttnBlocks}
    \Attn_g(X) = \frac{1}{1+g}
    \left(
    \underbrace{
    \begin{bmatrix} A_{AA}\,V_A \\ A_{BB}\,V_B \end{bmatrix}
    }_{\mathrm{intra}}
    + \; g
    \underbrace{
    \begin{bmatrix} A_{AB}\,V_B \\ A_{BA}\,V_A \end{bmatrix}
    }_{\mathrm{inter}}
    \right).
\end{align}
This construction satisfies three essential properties: (i) at $g=0$, the replicas are exactly decoupled; (ii) the coupling is perfectly symmetric under $A \leftrightarrow B$ exchange; and (iii) the weights $1/(1+g)$ and $g/(1+g)$ sum to $1$, the mixture is normalized, so the residual stream scale does not grow trivially with $g$. This ensures that the pretrained network operates safely within its learned dynamic range. For stability we keep $g$ in range $[0,1]$.

Equation \eqref{eqn:AttnBlocks} is our proposed architectural realization of symmetric coupling in a pretrained DiT. In this realization, the difference mode $v$ interaction is mediated by self-attention routing across the two replicas. Mathematically, however, it will appear as a perturbation to a special state that we will define below.

\subsection{Linearized Analysis of the Attention Difference}
\label{sec:LinAttnDiff}
Given the above DiT architecture, consider two identical per layer outputs of the attention mechanism  $H^{A}_{\ell,0}=H^B_{\ell,0}$. We will refer to these identical outputs as the symmetric state. We want to study the per layer response to a symmetry breaking perturbation $h_{\ell}$ of this symmetric state. We then consider the deformed output of the attention mechanism
\begin{align}
    H^A_{\ell} = H_{\ell},_0 + \frac{h_{\ell}}{\sqrt2},
    \qquad
    H^B_{\ell} = H_{\ell,0} - \frac{h_{\ell}}{\sqrt2}.
\end{align}
Expanding the value projection to first order around the symmetric state, we get
\begin{align}
    V^A_{\ell} &= V_{\ell,0} + \delta h_{\ell} + \mathcal O(\|h_{\ell}\|^2),\\
    V^B_{\ell} &= V_{\ell,0}, - \delta h_{\ell} + \mathcal O(\|h_{\ell}\|^2),
\end{align}
with $\delta h_{\ell}= J_{\ell,V} h_{\ell}$ for an effective Jacobian operator $J_V$ that absorbs the per layer normalization and value projection.

Similarly, the key and query dot products induce perturbations in the attention weight matrices that respect the replica exchange symmetry $A \leftrightarrow B$ (which maps $h_\ell \to -h_\ell$)
\begin{align}
\label{eqn:LogitPert1}
    A_{\ell,AA} &= A_{\ell,0} + \delta A_{\ell}^{(+)} + \mathcal O(\|h_{\ell}\|^2),
    \qquad
    A_{\ell,BB} = A_{\ell,0} - \delta A_{\ell}^{(+)} + \mathcal O(\|h_\ell\|^2),\\
    \label{eqn:LogitPert2}
    A_{\ell,AB} &= A_{\ell,0} + \delta A_{\ell}^{(-)} + \mathcal O(\|h_\ell\|^2),
    \qquad
    A_{\ell,BA} = A_{\ell,0} - \delta A_\ell^{(-)} + \mathcal O(\|h_\ell\|^2),
\end{align}
where $\delta A_\ell^{(+)}$ and $\delta A_\ell^{(-)}$ are linear in $h_\ell$. They encode the Jacobians of the blockwise softmax acting on the perturbed intra replica and inter replica logits, respectively. 

Substituting in \eqref{eqn:AttnBlocks}, to first order in $\|h_\ell\|$, the contribution of the coupled attention layer to the perturbation is 
\begin{align}
\label{eqn:AttnDelta}
\mathrm{Attn}_{g,\ell}^A - \mathrm{Attn}_{g,\ell}^B
&=
\frac{1-g}{1+g}\;
2\,A_{\ell,0}\,\delta h_\ell
\;+\;
\frac{1}{1+g}\;
2\bigl(\delta A_\ell^{(+)} + g\,\delta A_\ell^{(-)}\bigr)\,V_{\ell,0}
\;+\;
\mathcal{O}(\|h_{\ell}\|^2)\,,
\end{align}
see \appref{app:DerAttnDelta} for details.
This attention difference is an equation that decomposes the linear response into two mechanistically distinct first order pathways i.e. one through the value map and one through the attention kernel. 

The first term, $A_{\ell,0}\,\delta h_\ell$, is a value perturbation routed by the unperturbed attention kernel. The symmetric attention pattern $A_{\ell,0}$ is held fixed, while the perturbation perturbs the value vectors, and the resulting signal is transported across token positions. Consequently, we will occasionally call this the spatial routing term. In other words, this is the contribution from the value path with the attention map frozen. Its prefactor $(1-g)/(1+g)$ decreases with $g$ and vanishes at $g=1$, so strong inter replica coupling suppresses this channel completely.

The second term, $\bigl(\delta A_\ell^{(+)} + g\,\delta A_\ell^{(-)}\bigr)V_{\ell,0}$, is a perturbation of the attention kernel. Here the perturbation enters through the query key softmax pathway, changing the attention weights themselves, which are then applied to the background values $V_{\ell,0}$. This is the Jacobian contribution from the attention pattern modulation channel. Unlike the routing term, it is suppressed only by $1/(1+g)$ and therefore remains nonzero even at $g=1$. Thus increasing $g$ shifts the leading linear response from a fixed kernel routing of a perturbed value signal towards a perturbation of the attention kernel itself.

\subsection{The Linearized Propagator}
\label{sec:propagator}
To further study the contribution of the coupled attention layer to the attention difference \eqref{eqn:AttnDelta},  we now introduce the per layer linear operators  
\begin{align}
\label{eqn:R-def}
     R_{\ell} = 2\,A_{\ell,0}\,J_{\ell,V}\,, \quad
    P_{\ell}(g)\,h_{\ell} = 2\bigl(\delta A^{(+)}_{\ell}
    + g\,\delta A^{(-)}_{\ell}\bigr)\,V_{\ell,0}\,,
\end{align}
and the gating functions
\begin{align}
\label{eqn:rho-xi}
    \rho(g) = \frac{1-g}{1+g}\,,\qquad
    \xi(g) = \frac{1}{1+g}\,,
\end{align}
so that the attention difference \eqref{eqn:AttnDelta} becomes
\begin{align}
\label{eqn:delta-attn-compact}
    \Delta\Attn_{g,\ell} = \rho(g)\,R_{\ell}\,h_\ell
    + \xi(g)\,P_{\ell}(g)\,h_\ell
    + \mathcal{O}(\norm{h_\ell}^2)\,.
\end{align}
The sequential residual structure of the DiT block \eqref{eqn:BlockUp1}--\eqref{eqn:BlockUp2} means that the MLP acts on the post attention hidden state $\tilde{H}_{\ell}$, not on the original input $H_{\ell}$. Denoting the attention contribution to the perturbation mode $v$ by
\begin{align}
\Delta\Attn_{g,\ell} = [\rho(g)\,R_{\ell} + \xi(g)\,P_{\ell}(g)]\,h_{\ell}
+ \mathcal{O}(\|h_{\ell}\|^2)
\end{align}
as in~\eqref{eqn:delta-attn-compact}, the MLP sees a shifted input whose difference component is $h_{\ell} + \Delta\Attn_{g,\ell}$. Linearizing the MLP around the symmetric state, the full one block difference mode update is therefore
\begin{align}
    h_{\ell}^{+}
    &= (\mathbb{I}_{\ell} + J^{\MLP}_{\ell,0})
    \bigl(\mathbb{I_{\ell}} + \rho(g)\,R_{\ell} + \xi(g)\,P_{\ell}(g)\bigr)\,h_{\ell}
    + \mathcal{O}(\|h_{\ell}\|^2)\,,
    \label{eqn:Kg-exact}
\end{align}
where $J_{0,\MLP}$ is the pointwise MLP Jacobian evaluated at the symmetric state, incorporating the adaptive gating vector $\beta_{\ell}$ and $\mathbb{I}$ is the per-layer identity matrix. Algebraically expanding the product yields
\begin{align}
\label{eqn:Kg}
    h^{+}_{\ell} &= K_g\,h_{\ell} + \mathcal{O}(\|h_{\ell}\|^2)\,,\quad \\
    \label{eqn:KgDef}
    K_g &= \mathbb{I}_{\ell} + J_{\ell,0}^{\MLP}
    + \rho(g)\,R_{\ell} + \xi(g)\,P_\ell(g)
    + J_{\ell,0}^{\MLP}\bigl[\rho(g)\,R_\ell
    + \xi(g)\,P_\ell(g)\bigr]\,.
\end{align}
The cross term arises because the MLP processes the attention updated residual stream.  However, it is second order in the residual gating scale. In DiT, the attention and MLP residual branches are gated by learned vectors $\alpha_{\ell}$ and $\beta_{\ell}$ (cf.~\eqref{eqn:BlockUp1}--\eqref{eqn:BlockUp2}), which are
initialized at zero and remain $\mathcal{O}(\epsilon)$ throughout training. Since $J_{\ell,0}^{\MLP} = \mathcal{O}(\beta_{\ell})
= \mathcal{O}(\epsilon)$ and $R_\ell, P_\ell(g) = \mathcal{O}(\alpha_{\ell}) = \mathcal{O}(\epsilon)$, the cross term scales as $\mathcal{O}(\epsilon^2)$ while the individual attention and MLP contributions scale as $\mathcal{O}(\epsilon)$. We therefore approximate
\begin{align}
\label{eqn:Kg-approx}
    K_g \approx \mathbb{I_\ell} + J^{\MLP}_{\ell,0}
    + \rho(g)\,R_\ell + \xi(g)\,P_\ell(g)\,,
\end{align}
and we have dropped terms of order $\mathcal{O}(\epsilon^2)$.

Around the unperturbed symmetric state $v = 0$, the LayerNorm Jacobian is well defined and is absorbed into the effective linear operators $R_{\ell}$, $P_{\ell}(g)$, and $J_{\ell,0}^{\MLP}$ without
altering the structure of the expansion \eqref{eqn:AttnDelta}.  Similarly, the adaptive residual gating vectors $\alpha_{\ell}$ and $\beta_{\ell}$ act as layer dependent, elementwise rescalings of the attention and MLP contributions, respectively. These vectors are absorbed into $R_{\ell}$, $P_{\ell}(g)$ (via $\alpha_{\ell}$) and $J_{\ell,0}^{\MLP}$ (via $\beta_{\ell}$).  Because these gating vectors are predicted from the conditioning embedding and are initialized to zero during training, they can introduce significant layer to layer variation in the effective propagator, which is one mechanism behind the depth localization.  Consequently, $K_g$ in~\eqref{eqn:Kg-approx} is the effective per layer Jacobian of the difference mode update around the symmetric state, accurate to leading order in the gating scale.

In this approximation, the MLP contribution $J_{\ell,0}^{\MLP}$ acts independently on each spatial token $\MLP(H)_n = \phi(H_n)$ for $n = 1,\ldots,N$, where $\phi$ is the same nonlinear function applied at every position. Therefore $J_{\ell,0}^{\MLP}$ is block diagonal in token space and cannot perform explicit spatial routing, it processes the difference mode token by token without cross position information exchange.  The dropped cross-term
$J_{\ell,0}^{\MLP}[\rho(g)\,R_\ell + \xi(g)\,P_\ell(g)]$ would inherit spatial routing structure from $R_\ell$, its omission is what preserves the clean separation between spatially routing and pointwise contributions in the propagator decomposition.

The linearized propagator $K_g$ \eqref{eqn:Kg} describes how the perturbation is transported through the network, but a linear map around a single fixed point cannot produce speciation, it can only damp or amplify the difference (cf. discussion of \
\S\ref{sec:coupled-reverse}). To derive a speciation criterion, we require a nonlinear score with multiple attractors. Therefore, we model the local marginal distribution of the perturbation at reverse step $s$ and layer $\ell$ using the simplest nontrivial case which is a symmetric two component Gaussian mixture \cite{biroli2024}\footnote{An extensions to multiple classes was introduced in \cite{achilli2026theory}.}
\begin{align}
\label{eqn:mixture}
    p_{s,\ell}(h) = \tfrac{1}{2}\,
    \mathcal{N}\!\bigl(h;\,+m_{s,\ell},\,C_{s,\ell}\bigr)
    + \tfrac{1}{2}\,
    \mathcal{N}\!\bigl(h;\,-m_{s,\ell},\,C_{s,\ell}\bigr)\,,
\end{align}
where $m_{s,\ell} \in \R^D$ is the branch separation vector and $C_{s,\ell} \succ 0$ is the local strictly positive definite covariance matrix. For bookkeeping purposes, we will call objects at fixed step $s$ and layer $\ell$, a local object. The corresponding score is
\begin{equation}\label{eq:mixture-score}
    \nabla_h \log p_{s,\ell}(h)
    = -C_{s,\ell}^{-1}\,h
    + C_{s,\ell}^{-1}\,m_{s,\ell}\,
    \tanh\!\bigl(m_{s,\ell}^\top C_{s,\ell}^{-1}\,h\bigr)\,.
\end{equation}
The first term is the linear restoring force, identical to
the single Gaussian case. The second term provides the nonlinearity, the $\tanh$ saturates for large
$|m^\top C^{-1} h|$, creating a cubic like bifurcation
structure that admits nontrivial fixed points.

We model the deterministic one block update by combining the
linearized propagator $K_g$ \eqref{eqn:Kg} with a local score gain
$\gamma_{s,\ell} > 0$ which is the discrete analogue
of $\beta_t$ times the inverse noise variance in the continuous
theory. In the pretrained DiT, it absorbs the combined effect of
the AdaLN modulation and the learned output
projection scale of each block. It is in principle measurable
from network activations by comparing the magnitude of the
score like update to the residual stream, though we do not
estimate it directly in the present experiments since the qualitative predictions such as the mode ordering, gap
existence, and gap collapse depend only on the
monotonicity and sign of $\gamma_{s,\ell}$, we leave the precise measurement of $\gamma_{s,\ell}$ to future work. In total, we obtain
\begin{align}
\label{eqn:one-block-map}
    h^{+} = K_g\,h + \gamma_{s,\ell}\Bigl[
    -C_{s,\ell}^{-1}\,h
    + C_{s,\ell}^{-1}\,m_{s,\ell}\,
    \tanh\!\bigl(m_{s,\ell}^\top C_{s,\ell}^{-1}\,h\bigr)
    \Bigr]\,.
\end{align}

The one block map \eqref{eqn:one-block-map} combines two
conceptually distinct contributions to the difference mode $v$
dynamics.  The propagator \eqref{eqn:Kg} captures the
architectural response of the block, how self-attention
routes the difference signal across spatial positions and how
the pointwise MLP reshapes it.  Because $K_g$ is obtained by
linearizing the block around the replica symmetric point
$v = 0$, where the mixture score \eqref{eq:mixture-score}
itself vanishes, $K_g$ encodes the Jacobian of the block
evaluated at the point of zero score.

The second term, proportional to $\gamma_{s,\ell}$, models
the effective per block score increment, the fraction
of the total score response attributable to block $\ell$ at
reverse step $s$, evaluated on the two component
mixture \eqref{eqn:mixture}.  In a deep residual network
with $L$~blocks, the full score is built up incrementally,
$\gamma_{s,\ell}$ parameterizes each block's share.

A potential concern is that the linearized part of the score
overlaps with the propagator $K_g$.  To see this
explicitly, expand~\eqref{eq:mixture-score} around the difference mode $v = 0$
\begin{align}
\label{eqn:score-expansion}
    \nabla_v \log p_{s,\ell}(v)
    &= -\Lambda_{\mathrm{eff}}\,v
    + \mathcal{O}(\|v\|^3)\,,
\end{align}
where effective precision of the mixture is
\begin{align}
\label{eqn:Lambda-eff}
    \Lambda_{\mathrm{eff}}
    = C_{s,\ell}^{-1}
    - C_{s,\ell}^{-1}\,m_{s,\ell}\,m_{s,\ell}^\top\,
    C_{s,\ell}^{-1}\,.
\end{align}
The first piece $-C_{s,\ell}^{-1}\,v$ is the single Gaussian
restoring force and the rank one correction
$C_{s,\ell}^{-1}\,m_{s,\ell}\,m_{s,\ell}^\top\,
C_{s,\ell}^{-1}\,v$ is the linear contribution from expanding
the $\tanh$ and encodes the partial cancellation of the
restoring force along the branch separation direction.

One could therefore absorb all linear in $v$ score
contributions into a redefined propagator
\begin{align}
\label{eqn:Ktilde}
    \tilde K_g
    = K_g - \gamma_{s,\ell}\,\Lambda_{\mathrm{eff}}\,,
\end{align}
leaving only the genuinely nonlinear remainder
\begin{align}
\label{eqn:nonlinear-remainder}
    \gamma_{s,\ell}\,C_{s,\ell}^{-1}\,m_{s,\ell}
    \Bigl[
        \tanh\!\bigl(m_{s,\ell}^\top C_{s,\ell}^{-1}\,v\bigr)
        - m_{s,\ell}^\top C_{s,\ell}^{-1}\,v
    \Bigr]
    = \mathcal{O}(\|v\|^3)\,,
\end{align}
as the only score like term beyond the propagator.
This repartitioning leaves the fixed point equation \eqref{eqn:fixed-point-vector} and the scalar self consistency condition~\eqref{eqn:scalar-self-consistency} exactly invariant. Both conditions are derived from the full $\tanh$ nonlinearity, not from the linearization, so they already correctly account for the rank one linear piece.

This decomposition is analogous to separating the free propagator from the self energy in a diagrammatic expansion in quantum field theory \cite{zinn2021quantum}. In that context, $K_g$ plays the role of the bare propagator, $-\gamma_{s,\ell}\,\Lambda_{\mathrm{eff}}$ is the one loop self energy correction from the data distribution, and the $\tanh$ remainder \eqref{eqn:nonlinear-remainder} contains
the higher order vertices.  We treat $\gamma_{s,\ell}$ as an effective coupling, it absorbs any overlap between the architectural and score like linear contributions, so that the physically meaningful and measurable quantity is the total effective Jacobian at the symmetric point, $K_g - \gamma_{s,\ell}\,\Lambda_{\mathrm{eff}}$, not the individual partition.  All testable predictions depend on this total Jacobian and on the bifurcation structure of the $\tanh$ term, they are therefore invariant under repartitioning between~$K_g$ and~$\gamma_{s,\ell}$.

A fixed point satisfies
\begin{align}
\label{eqn:fixed-point-vector}
    \Bigl[(\mathbb{I} - K_g) + \gamma_{s,\ell}\,C_{s,\ell}^{-1}
    \Bigr]\,v
    = \gamma_{s,\ell}\,C_{s,\ell}^{-1}\,m_{s,\ell}\,
    \tanh\!\bigl(m_{s,\ell}^\top C_{s,\ell}^{-1}\,v\bigr)\,.
\end{align}
This is the discrete, attention gated analogue of the continuous fixed point equation in the symmetric coupled OU analysis of \cite{albrychiewicz2026dynamical}.  We emphasize
that the condition $v^{+} = v$ should not be read as a literal dynamical equilibrium of a single transformer block. Here, the difference mode flows through all blocks and reverse steps without equilibrating at any one of them.  Rather,
equation \eqref{eqn:fixed-point-vector} is a local stability criterion, at each $(s,\ell)$ it asks whether the effective one block potential for $v$ has developed a bifurcation point. This bifurcation point is where the symmetric solution $v=0$ becomes locally unstable and two new attracting directions appear, signaling the onset
of branch formation.  This is the standard interpretation of the self consistency equation in mean field theory \cite{mezard2009information} of phase transitions (cf., the Curie--Weiss model \cite{kittel2018introduction}), applied here to
the discrete computational graph of the transformer.

\subsection{Modewise Signal-to-Noise Ratio Formula}
\label{sec:snr}

We now present a mean field theory to determine the condition for the bifurcation point for each layer $\ell$ depending on the coupling strength $g$. We start with the projection onto fixed empirical modes and determine their bifurcation. Let
\begin{align}
C^{\mathrm{emp}}_{0,\ell}=\frac1M V_{0,\ell}^\top V_{0,\ell}\in\mathbb R^{D\times D}
\end{align}
denote the empirical covariance of the hidden state difference vectors at the initial reverse step $s=0$ ($t = T$) and layer $\ell$, where $V_{0,\ell}\in\mathbb R^{M\times D}$ stacks the $M$ sampled difference vectors row wise and $D$ is the dimension of each difference vector at layer $\ell$.  Let $\{r_k^{(\ell)}\}_{k=1}^K$ be its leading eigenvectors, obtained numerically through the dual Gram matrix
\begin{align}
\label{eqn:Gram}
G_{0,\ell}=\frac1M V_{0,\ell}V_{0,\ell}^\top\in\mathbb R^{M\times M}
\end{align}
via the Nystr\"om construction. For each reverse step $s$ and layer $\ell$, define the modal projections
\begin{align}
\label{eqn:modal-projections}
    c_{k} = r_k^\top C_{s,\ell}\,r_k\,,\qquad
    m_{k} = r_k^\top m_{s,\ell}\,,\qquad
    \eta_{k}(g) = r_k^\top K_g\,r_k\,,
\end{align}
here $c_k$ is a projected covariance, and $\eta_k(g)$ is a projected one block gain and we suppress the $(s,\ell)$ subscripts at left hand side for clarity.

In a mean field approximation, we set $v = a_k\,r_k$ and derive a scalar self consistency equation for each mode independently.  This decoupling is justified under two conditions.  First, the empirical modes $\{r_k\}$ must be approximately orthogonal eigenvectors of the difference covariance $C_{s,\ell}$, which our empirical construction (\S\ref{sec:Empirical}) enforces by design.  Second, the branch separation vector $m_{s,\ell}$ in the mixture model \eqref{eqn:mixture} must be approximately aligned with the principal subspace spanned by $\{r_k\}$. Under the single mode ansatz, the argument of the $\tanh$ nonlinearity collapses to a scalar
\begin{align}
    m_{s,\ell}^\top C_{s,\ell}^{-1}\,v = a_k \frac{m_k}{c_k} = u_k\,,
\end{align}
where we define $u_k$ as the rescaled modal order parameter.  If $v$ instead contained a superposition of nonindependent modes, the $\tanh$ argument would become a sum $\sum_j u_j$, and the nonlinearity  would generate cross mode couplings. However, because the $\tanh$  expansion contains no even powers, these nonlinear cross mode  corrections enter strictly at $\mathcal{O}(\|v\|^3)$. Furthermore, any linear mode mixing arises only from the off diagonal elements of $K_g$ in the $\{r_k\}$ basis. In practice, since $\{r_k\}$ span the dominant variance directions at initialization, these off diagonal terms are small. Consequently, near the symmetric bifurcation point where $|u_k|$ is small, we can neglect these subleading corrections  and treat the onset of instability for the leading modes independently.

Using the empirical spectral decomposition, we expand $\eta_k(g)$ with \eqref{eqn:Kg}
\begin{align}
\label{eqn:eta-decomp}
    \eta_{k}(g)
    = 1 + \lambda^{\MLP}_{k}
    + \rho(g)\,\chi_{k}
    + \xi(g)\,\pi_{k}\,,
\end{align}
where
\begin{align}
\label{eqn:modal-components}
    \lambda^{\MLP}_{k} = r_k^\top J_{\MLP}\,r_k\,,\qquad
    \chi_{k} = r_k^\top R\,r_k\,,\qquad
    \pi_{k} = r_k^\top P_g\,r_k\,.
\end{align}
Here $\lambda_k^{\MLP}$ is the pointwise MLP contribution, $\chi_k$ is the modewise spatial routing gain, the quantity that is frequency selective by virtue of the learned attention patterns $A_0$ and $\pi_k$ the pattern modulation correction. With the mode expansion and assuming that $C_{s,l}$ is diagonal in $r_k$ basis, we can rewrite fixed point equation as \eqref{eqn:fixed-point-vector} 
\begin{align}
    \Bigl[(1 - \eta_k(g)) + \frac{\gamma_{s,\ell}}{c_k}
    \Bigr]\,a_k
    = \frac{\gamma_{s,\ell}\,m_k}{c_k}\,
    \tanh\!\Bigl(\frac{m_k}{c_k}\,a_k\Bigr)\,.
\end{align}
In terms of $u_k$ this simplifies to
\begin{align}
\label{eqn:scalar-self-consistency}
    u_k = \kappa_{v,k}(s,\ell;\,g)\;\tanh(u_k)\,,
\end{align}
with the modewise speciation parameter
\begin{align}
\label{eqn:kappa-main}
    \kappa_{v,k}(s,\ell;\,g)
    = \frac{\gamma_{s,\ell}\,m_k^2}
    {c_k\Bigl((1 - \eta_k(g))\,c_k + \gamma_{s,\ell}\Bigr)}\,.
\end{align}
Equation \eqref{eqn:scalar-self-consistency} has the standard mean field bifurcation structure for $\kappa_{v,k} \le 1$ the only solution is $u_k = 0$ when replicas are identical. For $\kappa_{v,k} > 1$ two additional nonzero solutions emerge symmetrically so replicas are committed to distinct branches. The modal speciation step is therefore defined by
\begin{align}
\label{eqn:speciation-criterion}
    \kappa_{v,k}\!\bigl(s_{\spec}^{(k)},\,\ell;\,g\bigr) = 1\,,
\end{align}
when the equation undergoes a standard pitchfork bifurcation.

Following the logic of \cite{albrychiewicz2026dynamical}, we factor the speciation parameter into a score gain and a signal-to-noise ratio
\begin{align}
\label{eqn:kappa-snr}
    \kappa_{v,k}(s,\ell;\,g)
    = \gamma_{s,\ell}\;\SNR_{v,k}(s,\ell;\,g)\,,
\end{align}
where the modewise difference mode SNR is
\begin{equation}
\label{eqn:snr-def}
    \SNR_{v,k}(s,\ell;\,g)
    = \frac{m_k^2}
    {c_k\Bigl((1 - \eta_k(g))\,c_k
    + \gamma_{s,\ell}\Bigr)}\,.
\end{equation}
Substituting the decomposition \eqref{eqn:eta-decomp} and writing $\mu_k = 1/c_k$ for the modal precision, this expands to
\begin{align}
\label{eqn:snr-expanded}
    \SNR_{v,k}(s,\ell;\,g)
    = \frac{m_k^2\,\mu_k^2}
    { \gamma_{s,\ell}\mu_k-\lambda_k^{\MLP} - \rho(g)\,\chi_k
    - \xi(g)\,\pi_k \,}.
\end{align}
The SNR formula \eqref{eqn:snr-expanded} is well defined and
positive provided the denominator is strictly positive, i.e.,
\begin{align}
\label{eqn:snr-positivity}
    \gamma_{s,\ell}\,\mu_k
    > \lambda_k^{\MLP}
    + \rho(g)\,\chi_k
    + \xi(g)\,\pi_k\,,
\end{align}
or equivalently $\eta_k(g) < 1 + \gamma_{s,\ell}/c_k$. This condition states that the combined score gain and modal precision must exceed the per block amplification of the difference mode.  In the opposite regime, $\eta_k(g) \ge 1 + \gamma_{s,\ell}/c_k$, the linearized difference mode is amplified faster than the score can restore it, and the mean field picture breaks down and a nonperturbative treatment would be required.

To obtain an analytic theory, we adopt the phenomenological ansatz that the branch separation amplitude propagates multiplicatively along the trajectory,
\begin{align}
\label{eqn:ModePropagator}
m_{k,s,\ell} = G_{v,k}(s,\ell;\,g)\,m_{k,\mathrm{init}}.
\end{align}
This relation is not derived from first principles in the present work, rather, it is an empirically testable closure consistent with the linearized propagation picture. $G_{v,k}$ is an effective cumulative gain
\begin{align}
\label{eqn:cumulative-gain}
    G_{v,k}(s,\ell;\,g)
    = \prod_{(s',\ell') \prec (s,\ell)}
    \eta_{k,s',\ell'}(g)\,,
\end{align}
where the ordered pair $(s',\ell')$ indexes the reverse step $s'$ and transformer layer~$\ell'$, and the relation $(s',\ell') \prec (s,\ell)$ means that the block at $(s',\ell')$ is processed earlier in the reverse trajectory than the block at $(s,\ell)$. Thus $G_{v,k}(s,\ell;\,g)$ accumulates the gain from the first block of the trajectory up to (but not including) block $(s,\ell)$. The propagated SNR becomes
\begin{equation}\label{eq:snr-propagated}
    \SNR_{v,k}(s,\ell;\,g)
    = \frac{G_{v,k}(s,\ell;\,g)^2\,m_{k,\mathrm{init}}^2}
    {c_{k,s,\ell}\Bigl((1 - \eta_{k,s,\ell}(g))\,c_{k,s,\ell}
    + \gamma_{s,\ell}\Bigr)}\,.
\end{equation}
This is the discrete counterpart of the continuous SNR formula. The OU processes eigenvalues are replaced by the learned attention routing gains $\chi_k$, and the coupling enters through $\rho(g)$, $\xi(g)$ and the cumulative discrete gain $G_{v,k}^2(g)$. Heuristically, one may view the layerwise propagation of the difference mode as resembling a discrete inverse renormalization group flow. An example of RG flow application to diffusion models can be found in \cite{masuki2025generative}. In this analogy, the reverse step $s$ and layer depth $\ell$ play the role of a discrete scale parameter, while the learned attention map $A_0$ acts as a data dependent spatial averaging operator. In this framework, the cumulative gain equation \eqref{eqn:ModePropagator} can be interpreted as representing the discrete flow of the modal operators, and the mode wise branch separation $m_{k}$ evolves according to the propagator $\eta_k(g)$. We stress, however, that in this paper, this is only an interpretive analogy, no exact RG identification is required for the derivations above. 

\subsection{The Synchronization Gap and Its Collapse}
\label{sec:gap-collapse}

We will now focus on deriving the central experimental observable that we are interested in i.e. the synchronization gap between global and local difference modes.

Let $k_{\mathrm{hi}}$ denote a leading mode associated with coarse, globally organized structure and \(k_{\mathrm{lo}}\) a trailing mode associated with finer detail. In the coherent low frequency regime characterized in \appref{app:RoutingDom}, we assume the following structural properties. First, the spatial routing contribution dominates the pattern-modulation correction,
\begin{align}
    |\pi_k| \ll |\chi_k|\,,
\end{align}
we provide a structural reason why this is valid in \appref{app:RoutingDom}.
Next, the MLP term is only weakly mode dependent as we discussed in \secref{sec:propagator}
\begin{align}
    \lambda_k^{\MLP} \approx \lambda^{\MLP}\quad\,.
\end{align}
Finally, as a modeling hypothesis consistent with the routing dominant regime, we assume that attention routes leading coarse modes more strongly than trailing fine modes,
\begin{align}
\chi_{k_{\mathrm{hi}}} > \chi_{k_{\mathrm{lo}}},
\end{align}
from RG perspective this can be interpreted as a coarse graining operator preserving $k_{\text{hi}}$ (long wavelength) while heavily suppresses $k_{\text{lo}}$ (respectively short wavelength). Under these structural properties the propagator \eqref{eqn:eta-decomp} simplifies to
\begin{align}
    \eta_k(g) \approx 1 + \lambda^{\MLP} + \rho(g)\,\chi_k\,,
\end{align}
and the routing dominant SNR becomes \eqref{eqn:snr-def}
\begin{align}
\label{eqn:snr-routing-dominant}
    \SNR_{v,k}(s,\ell;\,g)
    \approx \frac{m_k^2}
    {c_k\Bigl((-\lambda^{\MLP}- \rho(g)\,\chi_k)\,c_k
    + \gamma_{s,\ell}\Bigr)}\,.
\end{align}
If the remaining modal factors ($m_k$, $c_k$) are comparable, then $\SNR_{v,k_{\mathrm{hi}}} > \SNR_{v,k_{\mathrm{lo}}}$, which implies
\begin{equation}
    s_{\spec}^{(k_{\mathrm{hi}})}(\ell;\,g)
    < s_{\spec}^{(k_{\mathrm{lo}})}(\ell;\,g),
\end{equation}
the global mode speciates before the local mode. The modewise synchronization gap at layer~$\ell$ and coupling~$g$ is
\begin{align}
\label{eqn:sync-gap-def}
    \Delta s_v(\ell;\,g) :=
    s_{\spec}^{(k_{\mathrm{lo}})}(\ell;\,g)
    - s_{\spec}^{(k_{\mathrm{hi}})}(\ell;\,g)\,.
\end{align}
A positive value indicates that the global mode has committed while the local mode remains unresolved. Using the assumptions from above, the SNR difference between modes picks up a factor of $\rho(g)$ from the routing term. To first order in the spectral split
\begin{align}
\label{eqn:snr-diff-scaling}
    \SNR_{v,k_{\mathrm{hi}}}(s,\ell;\,g)
    - \SNR_{v,k_{\mathrm{lo}}}(s,\ell;\,g)
    = \mathcal{O}\!\left(\frac{1-g}{1+g}\right)\,,
\end{align}
hence the spatial routing contribution to the SNR difference vanishes as $g \to 1$. 

\subsection{Summary of Testable Predictions}
\label{sec:predictions-summary}
The derived theoretical framework yields four concrete empirical predictions. 
\begin{enumerate}
    \item \textbf{Commitment Hierarchy:}  Latent low frequency structures commit to a basin of attraction significantly earlier than high frequency textures ($\tau_{\mathrm{low}} < \tau_{\mathrm{high}}$). The speciation time \(\tau_{\mathrm{spec}}\) is expected to track this transition but is not required by the theory to lie strictly between them.
    \item \textbf{Depth Localization:}  Since the speciation time depends on the layer the synchronization gap will not be uniform across them.
    \item \textbf{Existence of the Natural Gap:} At $g=0$, fully decoupled replicas sharing an initialization $z_T$ exhibit an intrinsic synchronization gap whose magnitude and mode dependence are predicted by the routing dominant SNR formula~\eqref{eqn:snr-routing-dominant}.
    \item \textbf{Coupling Induced Collapse:} As symmetric coupling $g$ increases toward $1$, the spatial routing difference is suppressed, forcing both the internal layerwise gap and the behavioral commitment gap to collapse.
\end{enumerate}

\section{Empirical Setup}
\label{sec:Empirical}

In this section we describe experimental setup to verify theoretical predictions made in the previous section. 

We begin with variance preserving initialization to ensure that the difference mode $v$ measurement is not
confounded by pushing it out of prior distribution. We initialize the replicas with an antisymmetric perturbation that preserves the marginal variance of the generation prior probability
\begin{align}
    z_T \sim \mathcal{N}(0, \mathbb{I})\,,\quad
    \delta \sim \mathcal{N}(0, \mathbb{I})\,,\quad
    z_T^{A,B}
    = \frac{z_T \pm \sigma\,\delta}{\sqrt{1 + \sigma^2}}\,.
\end{align}
This guarantees
$\mathrm{Var}(z_T^A) = \mathrm{Var}(z_T^B) = \mathbb{I}$ while
injecting a controlled difference signal
\begin{align}
    v_T = \sqrt{2}\,\sigma\,\delta / \sqrt{1+\sigma^2}.
\end{align}
The inter replica correlation is
\begin{equation}
\label{eqn:correlation}
    \varrho(\sigma)
    = \frac{1 - \sigma^2}{1 + \sigma^2}\,,
\end{equation}
for the experimentally relevant range $0\le \sigma \le 1$, this interpolates from identical replicas $\varrho=1$ to independent replicas $\varrho=0$. For $\sigma>1$, the replicas become negatively correlated.

\subsection{Protocol I: Speciation Time and Scale Dependent Commitment}
\label{sec:EmpProtI}
Our first empirical protocol probes the speciation time \eqref{eqn:speciation-criterion} of the generative model. Two replicas are initialized with a shared macroscopic structure, a shared initial latent state $z_T$ and an antisymmetric noise perturbation, and coupled with strength $g$ up to an intervention step $t_{\mathrm{int}}$, after which they evolve independently $g=0$ using DDIM sampler \cite{song2020denoising} with $\eta=1$ effectively recovering DDPM \cite{ho2020denoising}. By measuring the divergence of the final decoded outputs $x^A(t_{\mathrm{int}}; g)$ and $x^B(t_{\mathrm{int}}; g)$, we establish at what time bifurcation described in the theory part occurs. 

We measure output agreement with two methods: within class perceptual speciation and scale dependent output commitment. To determine when the stochastic DDPM sampler commits the trajectories to the same semantic basin of attraction, we measure the
feature space cosine similarity of the final images using a pretrained ImageNet encoder (ResNet-50) \cite{he2016deep}.
\begin{align}
\label{eqn:cos-sim}
    a_{\mathrm{feat}}(t_{\mathrm{int}};g)
    =\cos\!\Bigl(\phi(x^A(t_{\mathrm{int}};g)),\, \phi(x^B(t_{\mathrm{int}};g))\Bigr).
\end{align}
Aggregating over $M$ paired seeds yields a median agreement curve. We estimate the speciation proxy $\tau_{\mathrm{spec}}(g)$ as the inflection point of a sigmoid fitted to this curve, providing an operational definition for macroscopic commitment.

To isolate how spatial routing resolves generative ambiguity, we decompose the image discrepancy into a coarse component and a residual fine component. Let $P$ denote adaptive average pooling to a fixed low resolution, and let $U$ denote bilinear upsampling back to the original image size. We then define
\begin{align}
\label{eqn:LowComit}
    d_{\mathrm{low}}(t_{\mathrm{int}};g)
    &= \|P(x_0^A)-P(x_0^B)\|_2^2,\\
\label{eqn:HighComit}
    d_{\mathrm{high}}(t_{\mathrm{int}};g)
    &= \| \bigl(x_0^A-U P(x_0^A)\bigr)
      - \bigl(x_0^B-U P(x_0^B)\bigr)\|_2^2.
\end{align}
Thus $d_{\mathrm{low}}$ measures discrepancy after strong spatial averaging, whereas $d_{\mathrm{high}}$ measures discrepancy in the corresponding high frequency residual. We extract the macroscopic $\tau_{\mathrm{g}}$ and microscopic $\tau_{\mathrm{l}}$ commitment times from the 50\% thresholds of their respective sigmoid fits. The output space synchronization gap is directly quantified as
\begin{align}
\label{eq:output-gap}
    \Delta\tau(g) = \tau_{\mathrm{l}}(g) - \tau_{\mathrm{g}}(g).
\end{align}
A positive gap formally indicates that the continuous physics of the DiT resolves low frequency global structure before high frequency local detail. In what follows, the terms global and local refer specifically to this output space scale decomposition, which is distinct from the internal hidden state basis defined in Protocol II.

\subsection{Protocol II: Internal Mode Stabilization and the Layerwise Gap}
\label{sec:EmpProtII}

Our second empirical protocol probes how the continuous synchronization gap is represented internally across the discrete Transformer depth $\ell$. In this setup, the replicas remain coupled at a constant $g$ for the entire reverse trajectory.

At each step $s$ and layer $\ell$, we extract the flattened hidden state difference vector for $M$ paired seeds
\begin{align}
    V_{s,\ell} = \frac{1}{\sqrt{2}} \begin{bmatrix} \mathrm{vec}(H_{s,\ell}^{A,(1)} - H_{s,\ell}^{B,(1)})^\top \\ \vdots \\ \mathrm{vec}(H_{s,\ell}^{A,(M)} - H_{s,\ell}^{B,(M)})^\top \end{bmatrix} \in \mathbb{R}^{M \times D}.
\end{align}
To construct a stable fixed basis of empirical principal modes without rank-deficiency artifacts, we form the \(M\times M\) dual Gram matrix \eqref{eqn:Gram} as discussed in \secref{sec:snr}. The normalized energy of mode $k$ over time is
\begin{align}
\label{eqn:mode-energy-fixed}
    \lambda_k(s,\ell) = \frac{\|V_{s,\ell} r_k^{(\ell)}\|^2}{\|V_{0,\ell} r_k^{(\ell)}\|^2}.
\end{align}
Equivalently,
\begin{align}
\lambda_k(s,\ell)=\frac{r_k^{(\ell)\top} C^{\mathrm{emp}}_{s,\ell} r_k^{(\ell)}}{\lambda_k^{(0,\ell)}},
\end{align}
where $\lambda_k^{(0,\ell)}$ is an eigenvalue of initial empirical covariance $C^{\mathrm{emp}}_{0,\ell}$.
If $r_k^{(\ell)}$ also diagonalizes $C^{\mathrm{emp}}_{s,\ell}$, these are eigenvalues of the current covariance otherwise these are fixed basis internal modes defined at $s=0$. They provide a stable internal coordinate system in which to ask how leading and trailing difference channel $v$ directions behave over time.

We then evaluate these normalized energies at the estimated speciation time from Protocol I for particular value of $g$
\begin{align}
    s = \tau_{\mathrm{spec}}(g).
\end{align}
For robustness, we aggregate clusters rather than relying on single indices
\begin{align}
\label{eqn:ProtIILead}
    \bar\lambda_{\mathrm{lead}}(\ell;g)
    &=
    \frac{1}{|\mathcal K_{\mathrm{lead}}|}
    \sum_{k\in \mathcal K_{\mathrm{lead}}}
    \lambda_k\!\bigl(\tau_{\mathrm{spec}}(g),\ell\bigr),\\
\label{eqn:ProtIITrail}
    \bar\lambda_{\mathrm{trail}}(\ell;g)
    &=
    \frac{1}{|\mathcal K_{\mathrm{trail}}|}
    \sum_{k\in \mathcal K_{\mathrm{trail}}}
    \lambda_k\!\bigl(\tau_{\mathrm{spec}}(g),\ell\bigr),
\end{align}
where $\mathcal K_{\mathrm{lead}}(\ell)$ and $\mathcal K_{\mathrm{trail}}(\ell)$
denote leading and trailing bands in the ordering induced by the initial covariance eigenvalues $\lambda_k^{(0,\ell)}$.
This yields the internal energy gap ratio
\begin{align}
\label{eq:internal-gap}
    \mathcal G_{\mathrm{int}}(\ell;g)
    =
    \frac{\bar\lambda_{\mathrm{trail}}(\ell;g)}
         {\bar\lambda_{\mathrm{lead}}(\ell;g)}.
\end{align}
Values $\mathcal G_{\mathrm{int}}(\ell;g) > 1$ indicate that trailing empirical modes retain more energy than leading empirical modes at the moment of the species commitment.

\section{Results}
\label{sec:Results}
In this section, we report the results of experiments described in the previous section. First, we determine the speciation time for a range of coupling strengths $g$ using the Protocol I experiment setup \secref{sec:EmpProtI}. Second, we conduct a layer wise sweep to analyze the synchronization gap and its sensitivity to g as defined in Protocol II\secref{sec:EmpProtII}.

\subsection{Protocol I Results}
As mentioned before, we use DDIM \cite{song2020denoising} reverse sampler for DiT-XL/2 model. For Protocol I experiment, we set the noise parameter $\eta=1$ which corresponds to stochastic DDPM sampler \cite{ho2020denoising}. We apply the coupling for all transformer blocks and we apply intra/inter replica blocks normalization as shown in \eqref{eqn:AttnBlocks}. For each $g$, we sweep $100$ reverse steps and we measure the cosine similarity of the final images in feature space using a pretrained ResNet-50 encoder. We use same ImageNet class id (0) in all experiments but we also verify that behavior does not change when we pick different class. We fit the sigmoid function to the median curve over $M=32$ seeds and define the speciation time as a midpoint, we also report the 95\% bootstrap confidence interval. We show results for $g\in\{0.1,0.3,0.5,0.7,0.9,1.0\}$ on \figref{fig:SpecG01}--\figref{fig:specG09} and also summarize them in \tabref{tab:SpecT} for readers' convenience.

\begin{figure}[ht]
    \centering
    \begin{subfigure}[b]{0.80\textwidth}\includegraphics[width=0.99\linewidth]{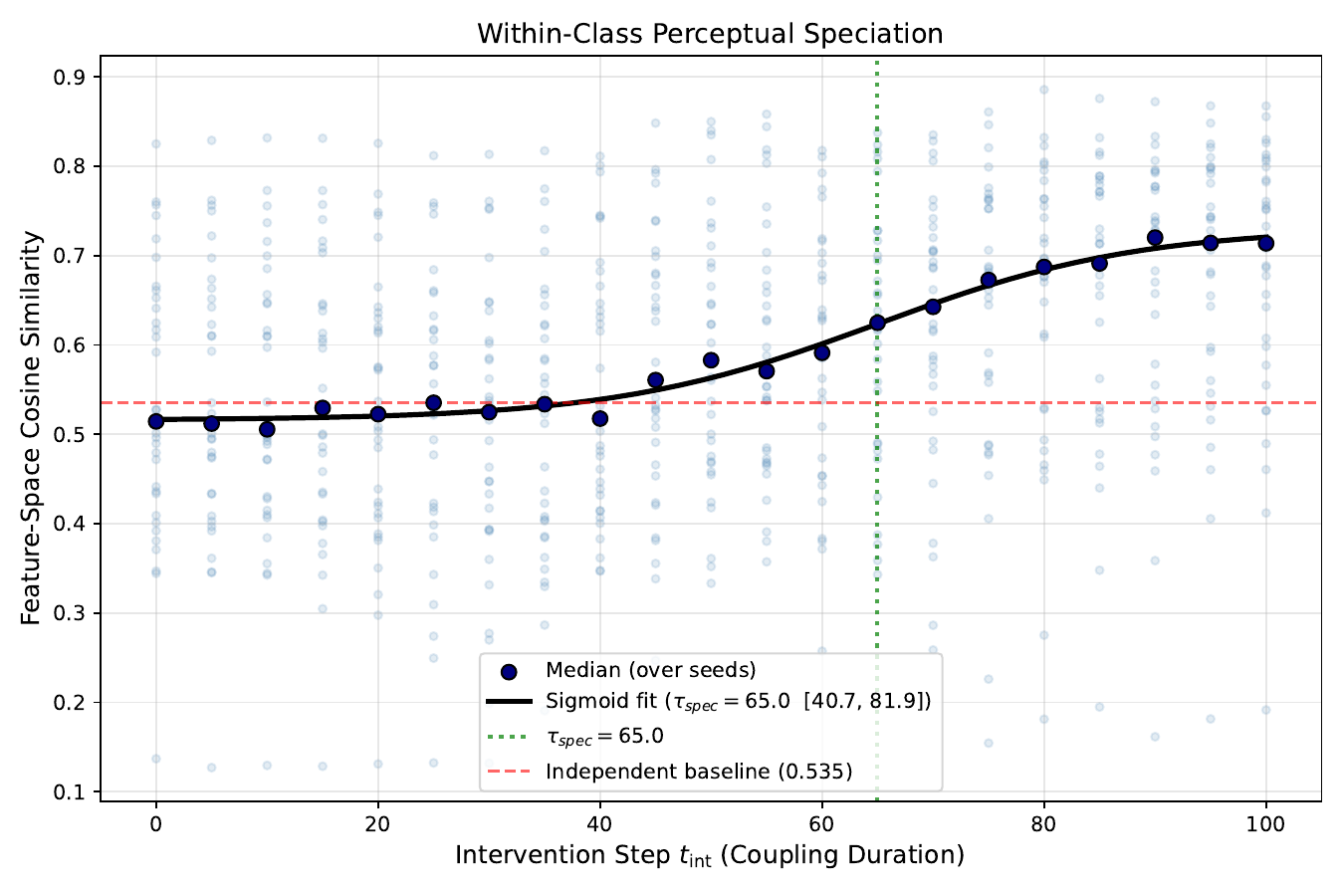}
    \caption{$g=0.1$}
    \label{fig:SpecG01}
    \end{subfigure}
    \begin{subfigure}[b]{0.80\textwidth}\includegraphics[width=0.99\linewidth]{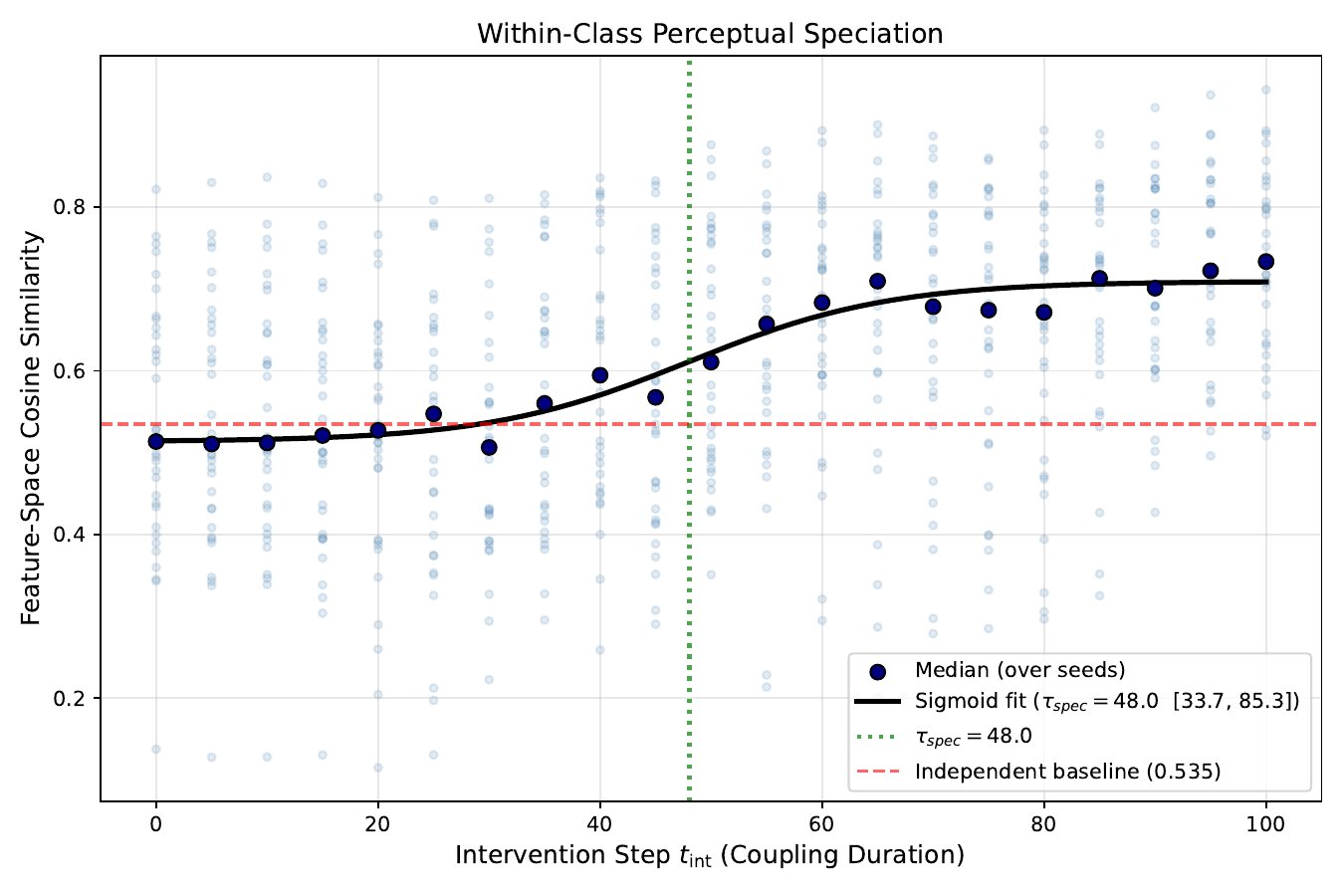}
    \caption{$g=0.3$}
    \end{subfigure}
    \caption{Each plot shows the median feature space cosine similarity \eqref{eqn:cos-sim} between paired outputs as a function of intervention step $t_{\mathrm{int}}$, together with the sigmoid fit used to extract $\tau_{\spec}(g)$. With an increase of coupling strength $g$, the speciation time decreases.}
\end{figure}
\begin{figure}
    \centering
    \ContinuedFloat
    \begin{subfigure}[b]{0.8\textwidth}\includegraphics[width=0.99\linewidth]{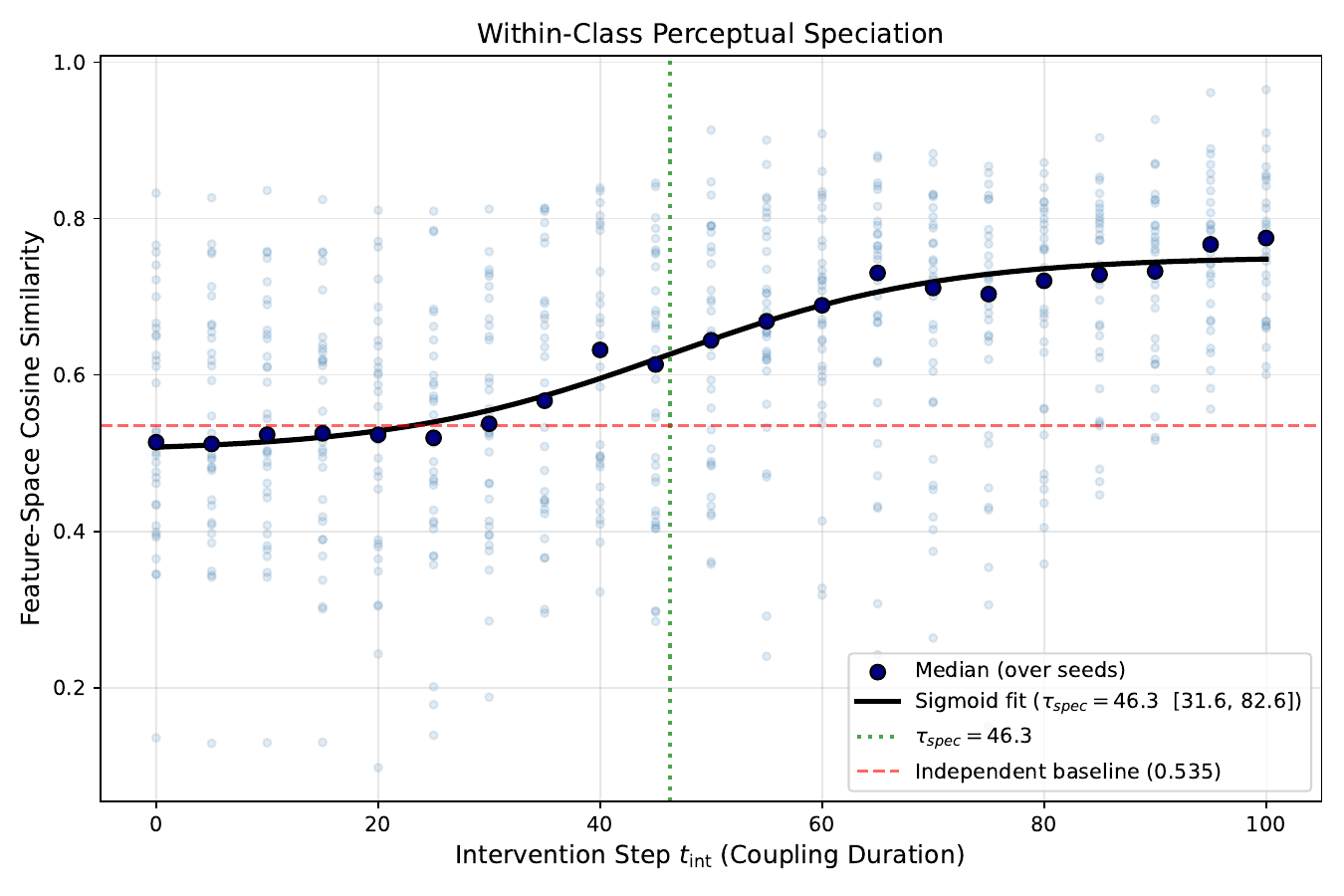}
    \caption{$g=0.5$}
    \end{subfigure}
    \begin{subfigure}[b]{0.8\textwidth}\includegraphics[width=0.99\linewidth]{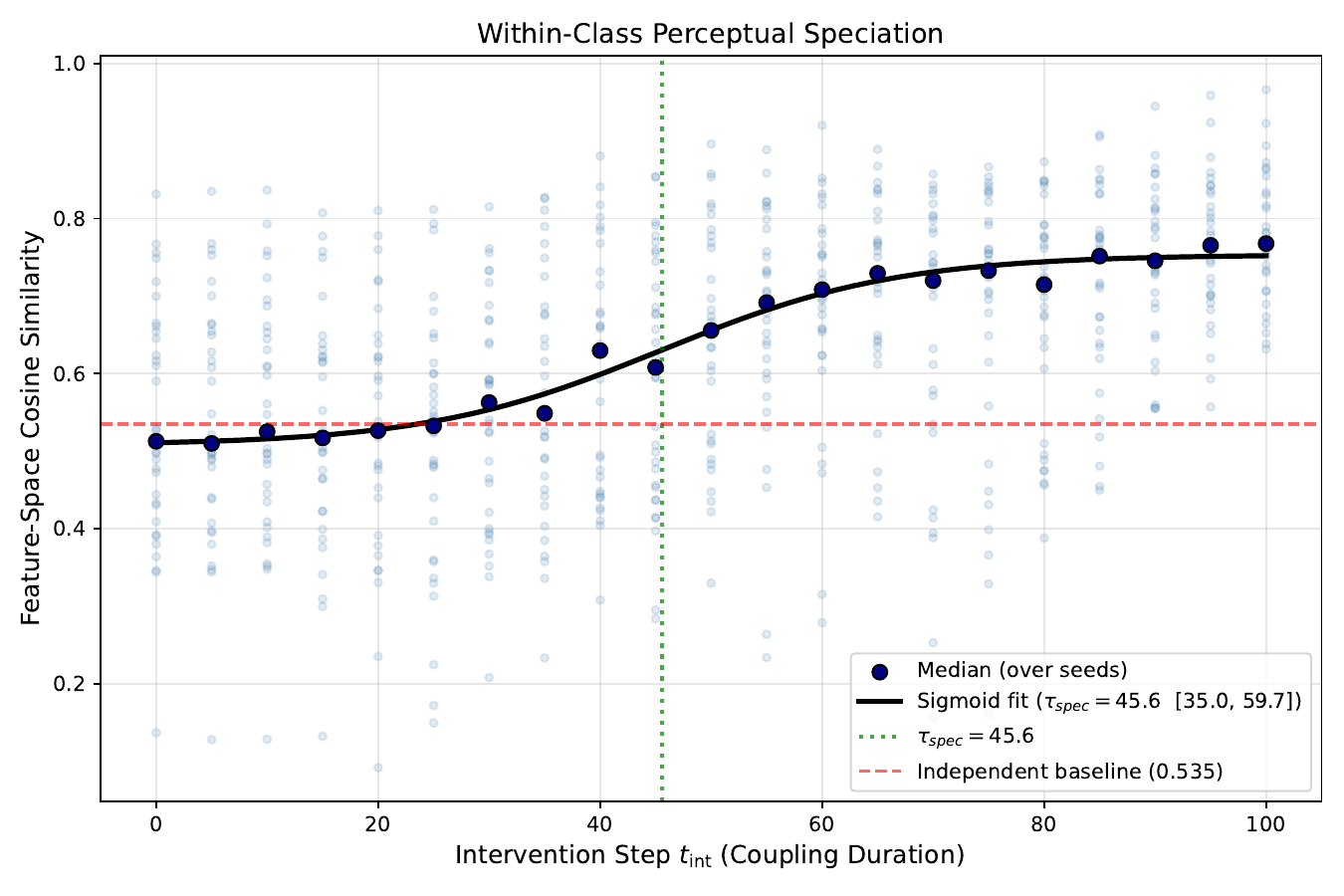}
    \caption{$g=0.7$}
    \end{subfigure}
     \caption{Each plot shows the median feature space cosine similarity \eqref{eqn:cos-sim} between paired outputs as a function of intervention step $t_{\mathrm{int}}$, together with the sigmoid fit used to extract $\tau_{\spec}(g)$. With an increase of coupling strength $g$, the speciation time decreases.}
\end{figure}
\begin{figure}
    \centering
    \ContinuedFloat
    \begin{subfigure}[b]{0.8\textwidth}\includegraphics[width=0.99\linewidth]{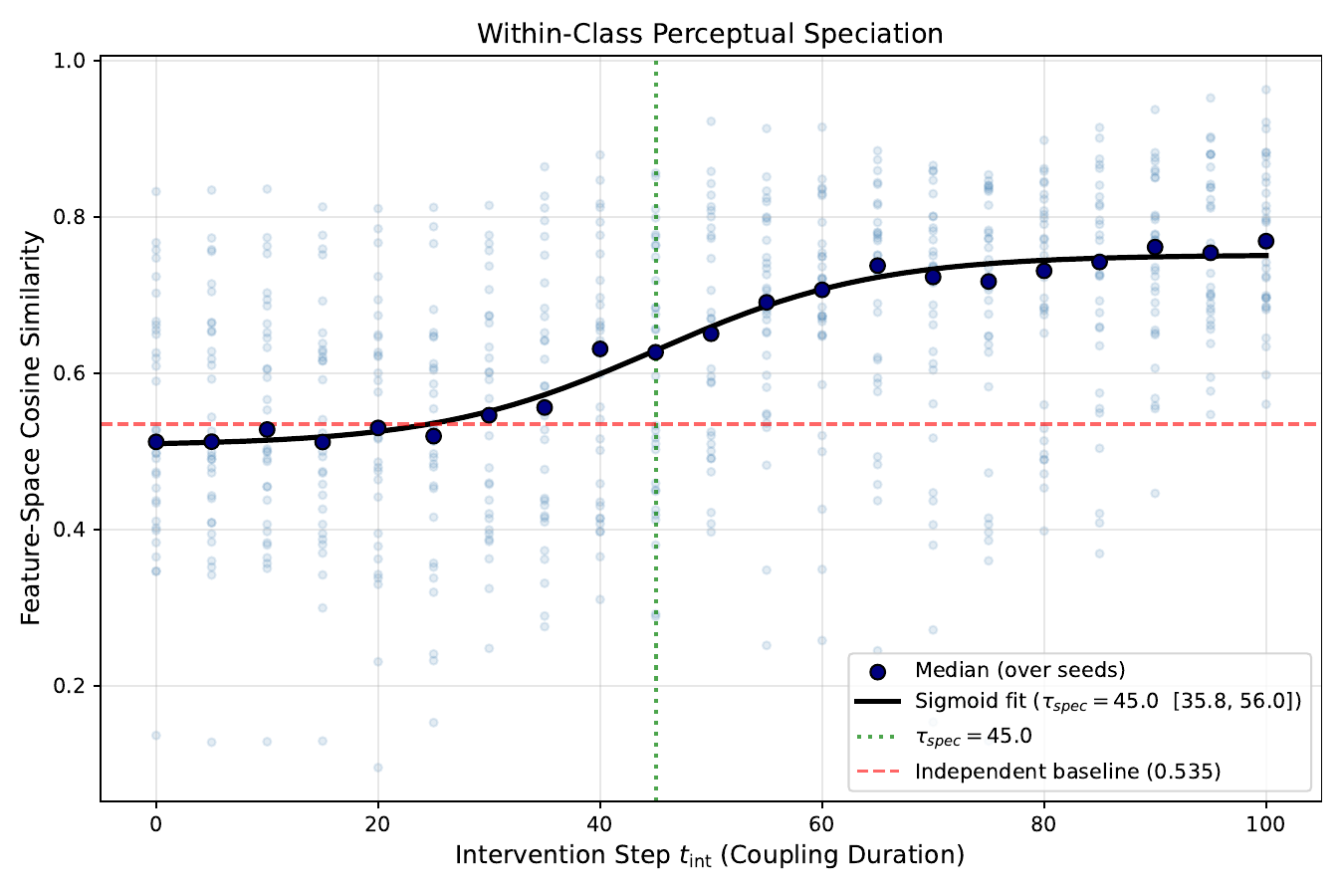}
    \caption{$g=0.9$}
    \end{subfigure}
    \begin{subfigure}[b]{0.8\textwidth}\includegraphics[width=0.99\linewidth]{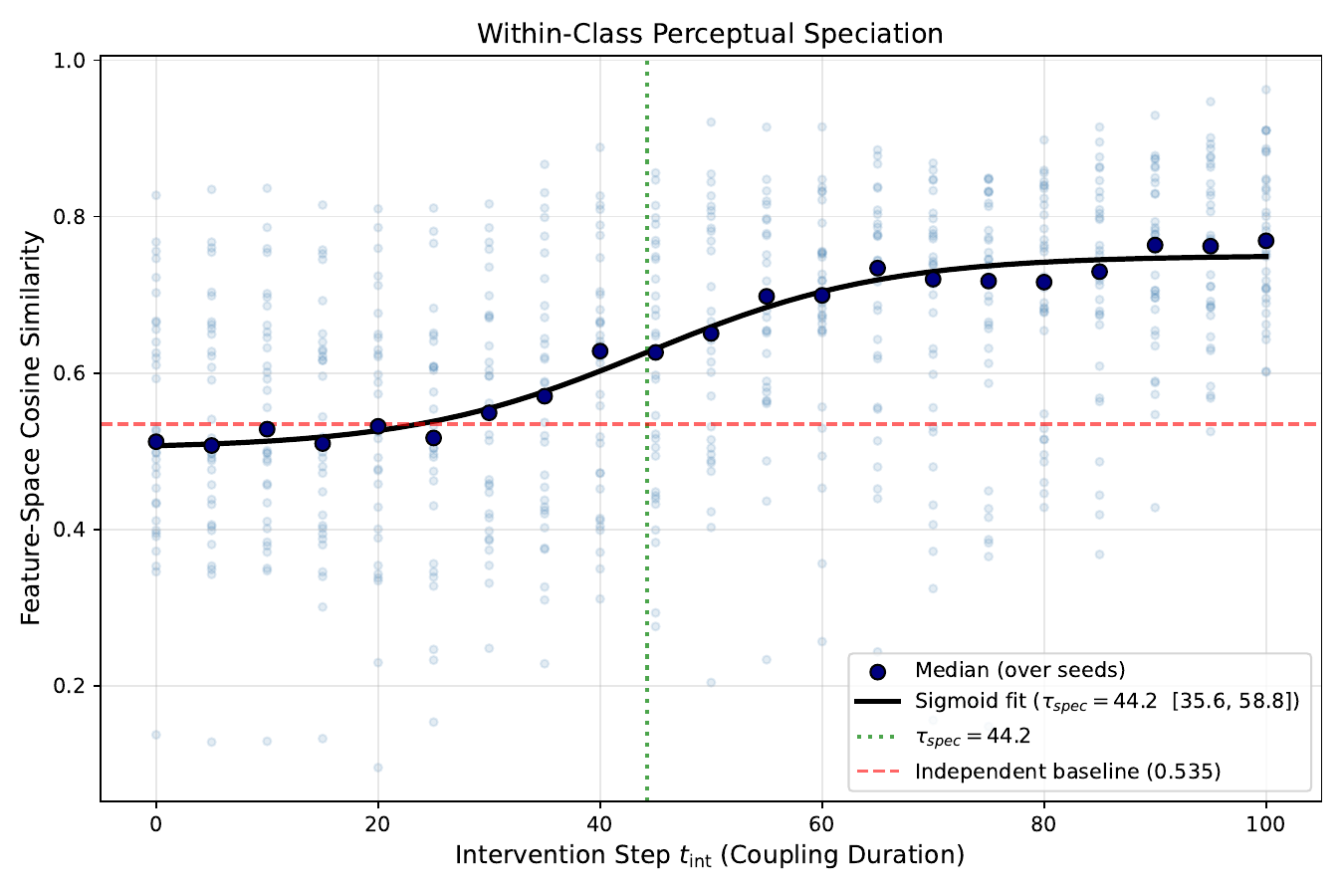}   
    \caption{$g=1.0$}
    \label{fig:specG09}
    \end{subfigure}
    \caption{Each plot shows the median feature space cosine similarity \eqref{eqn:cos-sim} between paired outputs as a function of intervention step $t_{\mathrm{int}}$, together with the sigmoid fit used to extract $\tau_{\spec}(g)$. With an increase of coupling strength $g$, the speciation time decreases.}
      \label{fig:spec3}
\end{figure}

\begin{table}[ht]
    \centering
    \begin{tabular}{|c| c|}
    \toprule
    Coupling strength $g$ & $\tau_{\spec}(g)$ [95\% CI] \\
    \midrule
    0.1 & $63.6\;[50.0,\,96.6]$ \\
    0.3 & $48.0\;[33.7,\,85.3]$ \\
    0.5 & $46.3\;[31.6,\,82.6]$ \\
    0.7 & $45.6\;[35.0,\,59.7]$ \\
    0.9 & $45.0\;[35.8,\,56.0]$ \\
    1.0 & $44.2\;[35.6,\,58.8]$ \\
\bottomrule
\end{tabular}
    \caption{Summary of speciation times determined using cosine similarity method and corresponding confidence intervals vs coupling strength $g$.}
    \label{tab:SpecT}
\end{table}

Qualitatively, all curves exhibit the same structure predicted by the theory for small $t_{\mathrm{int}}$, the paired trajectories remain close to the independently initialized baseline, while for sufficiently long shared evolution the feature space agreement rises and saturates at a substantially higher level. The main coupling effect is therefore not a change in the final plateau itself, but an earlier step shift of the transition. This is consistent with the interpretation that stronger symmetric coupling suppresses replica difference spatial routing earlier in the trajectory and thereby reduces the temporal window over which the two replicas can diverge before committing to the same semantic basin. 

\begin{figure}
    \centering
    \includegraphics[width=0.8\linewidth]{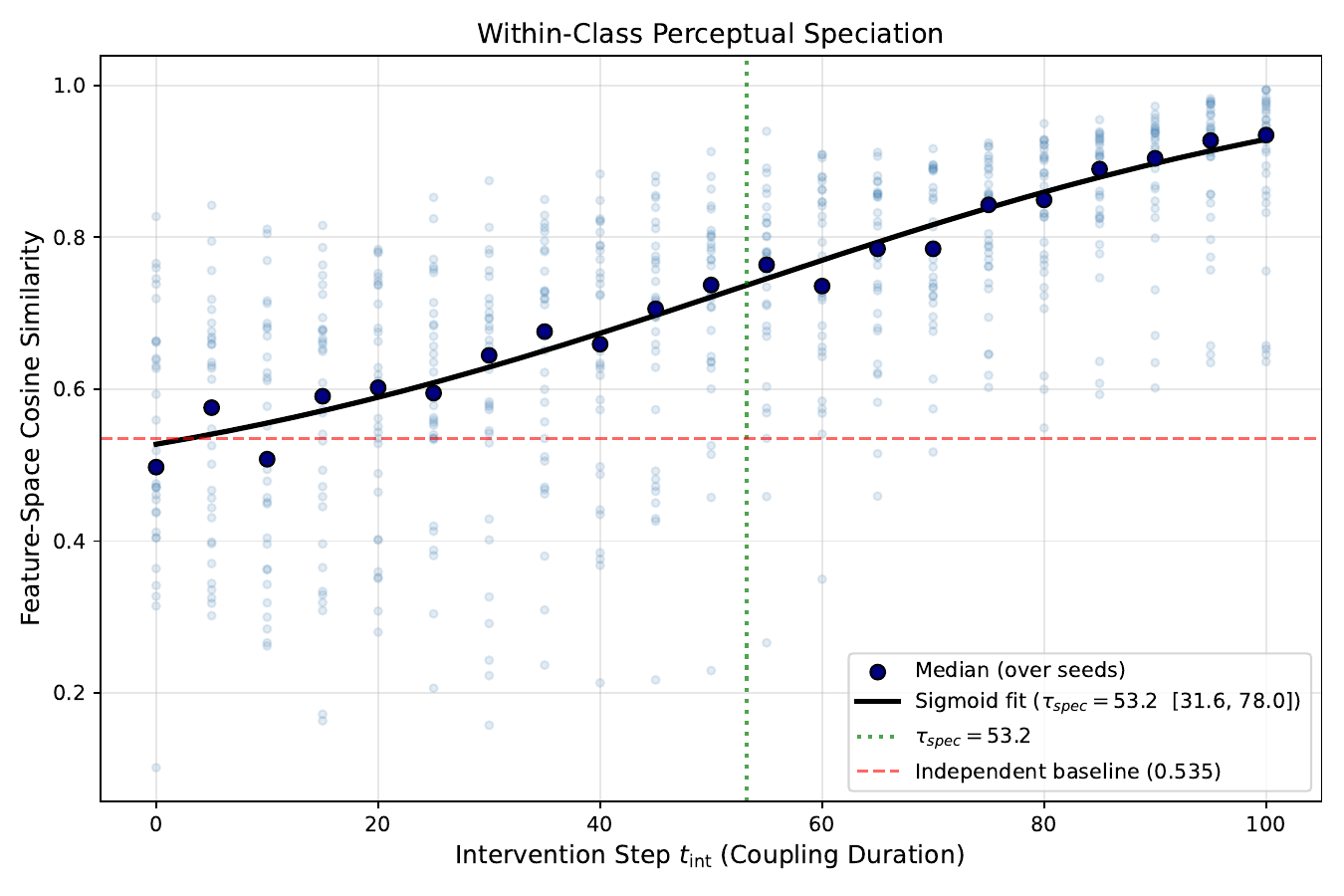}
    \caption{A separate analysis of the speciation time using the median feature space cosine similarity \eqref{eqn:cos-sim} between paired outputs for the vanishing coupling strength $g=0$. In this modified setup, the noise of reverse process for both replicas are shared until intervention time when it becomes independent.}
    \label{fig:SpecG0}
\end{figure}

We report only values of $g>0$ since the experiment is about measuring the intervention time when $g$ is switched to $0$. In the case of $g=0$ we can plot a separate uncoupled control where instead two replicas share noise until intervention time when noise is switched to be independent. For such setup we can again ask at what point these two trajectories finish at the same basin of the attraction and this is shown on \figref{fig:SpecG0}. In this case speciation time is earlier than for $g=0.1$ but this is due to the different intervention type as explained above. 

A second output of Protocol I is a measurement of a scale dependent output commitment. Here we use the coarse and fine commitment scores defined in \eqref{eqn:LowComit}, \eqref{eqn:HighComit} respectively, and fit separate sigmoids to obtain the midpoint times of global and local output alignment. For readability, we denote these fitted midpoints by $\tau_{\mathrm{g}}(g)$ and $\tau_{\mathrm{l}}(g)$, corresponding to the blue and red curves in the decoupling plots. As in the previous case we sweep coupling strength $g\in\{0.1,0.3,0.5,0.7,0.9,1.0\}$.

The measurements presented on \figref{fig:ComitG0}--\figref{fig:ComitG09} show that coarse output structure stabilizes substantially earlier than fine detail throughout the explored coupling range. Thus, the decoupling probe supports the existence of a  output space gap. This gap is extremely robust, after the weak coupling regime, the gap stabilizes near $\Delta\tau\approx 39\text{--}41$ steps across the medium and strong coupling regime. Importantly, these scale dependent midpoint times should not be identified with the primary speciation time $\tau_{\spec}(g)$. The decoupling probe measures when coarse and fine pixel space discrepancies become suppressed relative to an independent baseline. By construction, it is sensitive to the chosen low/high decomposition, the baseline, and residual variability. In contrast, $\tau_{\spec}(g)$ is defined from cosine similarity in the feature space of a pretrained ResNet-50 encoder and is intended to measure semantic agreement between the final paired samples, i.e., whether the two trajectories have committed to the same basin of attraction. For this reason, we use the cosine similarity as the primary estimator of speciation time and treat the decoupling probe as a complementary diagnostic. The former is closer to the theoretical notion of branch commitment. The latter is still essential, because it reveals the internal ordering of coarse versus fine output stabilization and thereby quantifies the observable synchronization gap in image space.

\begin{figure}[ht]
    \centering
    \begin{subfigure}[b]{0.8\textwidth}\includegraphics[width=0.99\linewidth]{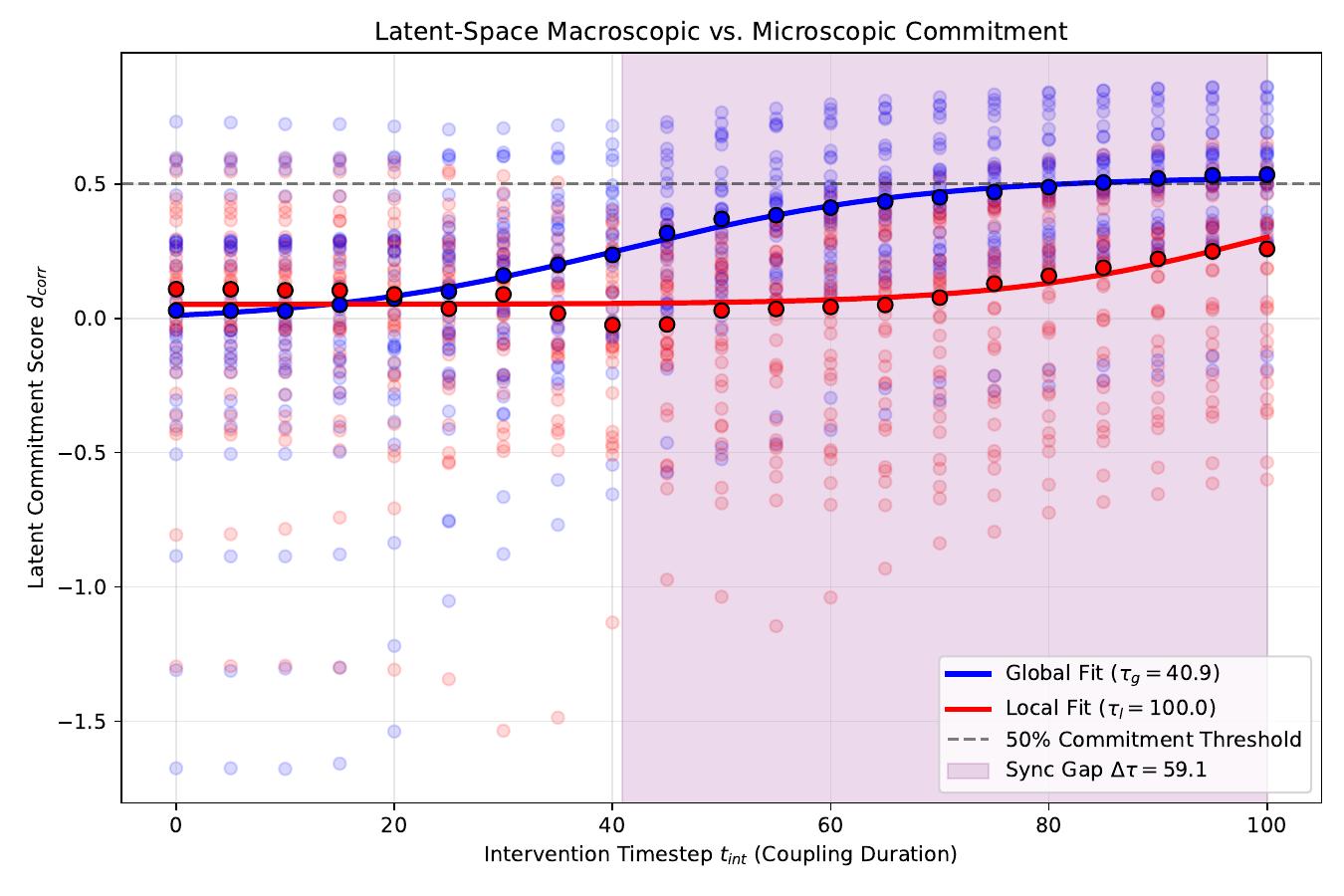}
    \caption{$g=0.1$}
    \label{fig:ComitG0}
    \end{subfigure}   
    \begin{subfigure}[b]{0.8\textwidth}\includegraphics[width=0.99\linewidth]{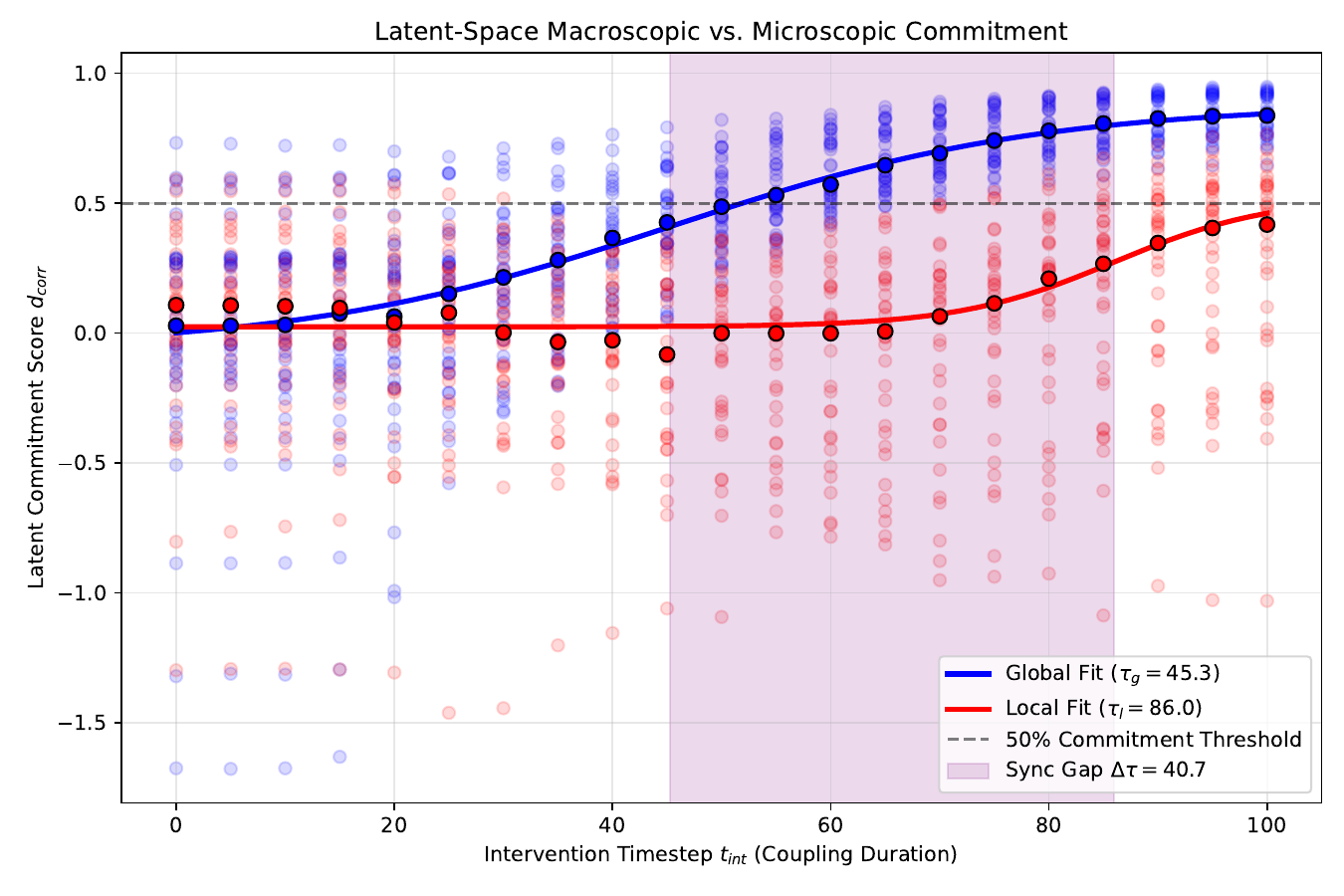}
    \caption{$g=0.3$}
    \end{subfigure}
    \caption{Each plot shows a global \eqref{eqn:LowComit} and local \eqref{eqn:HighComit} commitment scores vs coupling strength $g$. These show that coarse output structure stabilizes substantially earlier than fine detail throughout the explored coupling range.}
\end{figure}
\begin{figure}
    \centering
    \ContinuedFloat
    \begin{subfigure}[b]{0.8\textwidth}\includegraphics[width=0.99\linewidth]{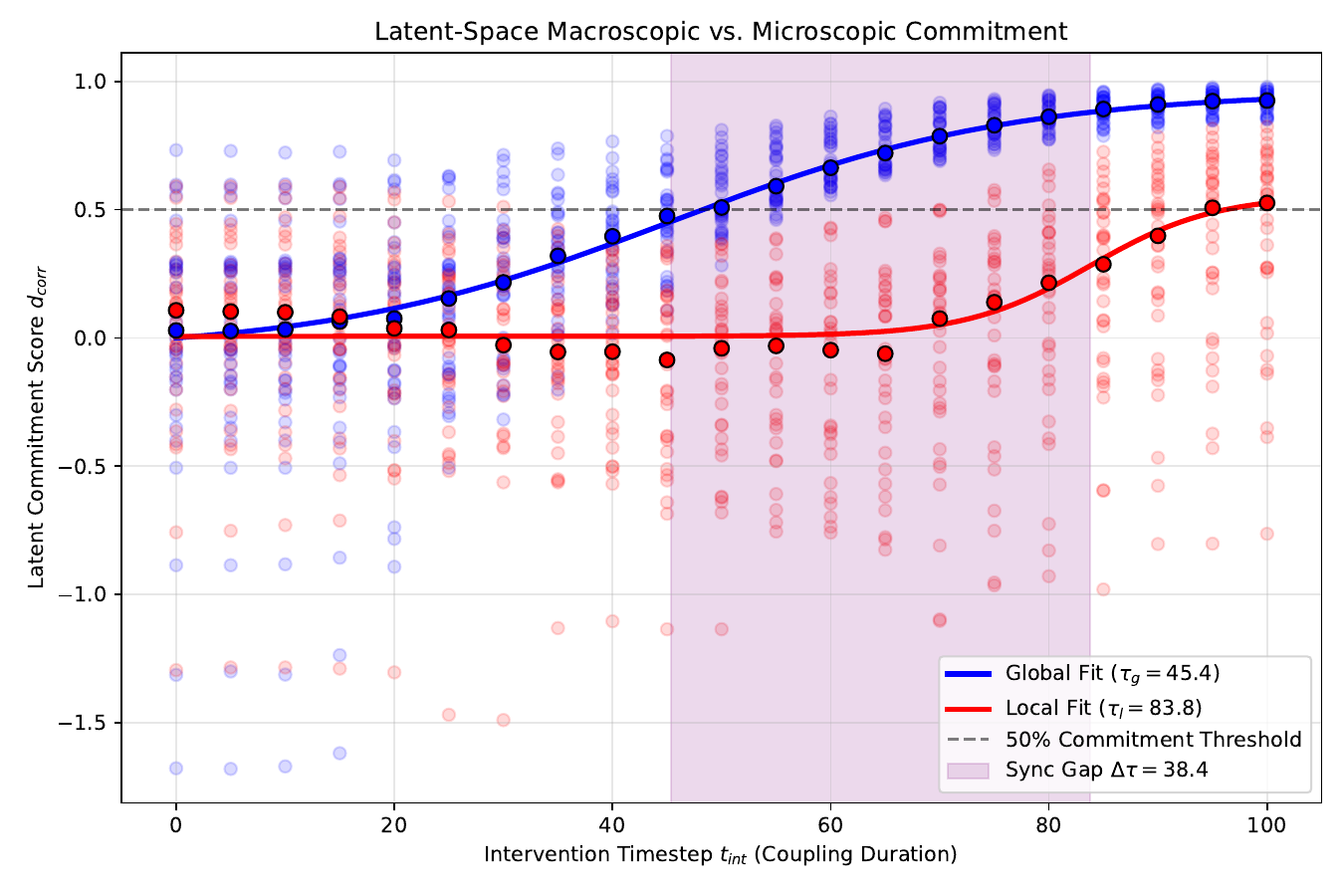}
    \caption{$g=0.5$}
    \end{subfigure}
    \begin{subfigure}[b]{0.8\textwidth}\includegraphics[width=0.99\linewidth]{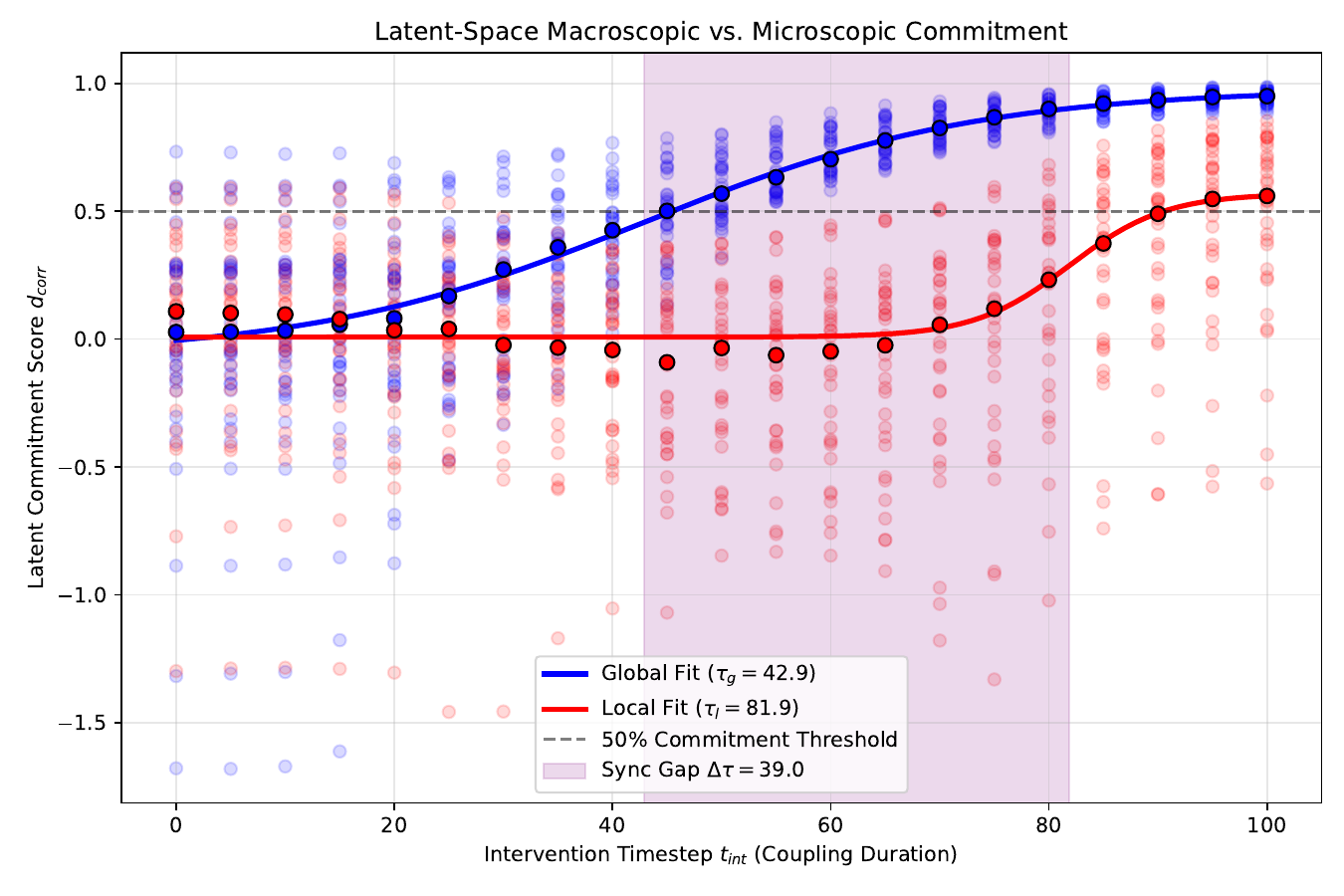}
    \caption{$g=0.7$}
    \end{subfigure}
    \caption{Each plot shows a global \eqref{eqn:LowComit} and local \eqref{eqn:HighComit} commitment scores vs coupling strength $g$. These show that coarse output structure stabilizes substantially earlier than fine detail throughout the explored coupling range.}
\end{figure}
\begin{figure}
    \centering
    \ContinuedFloat
    \begin{subfigure}[b]{0.8\textwidth}\includegraphics[width=0.99\linewidth]{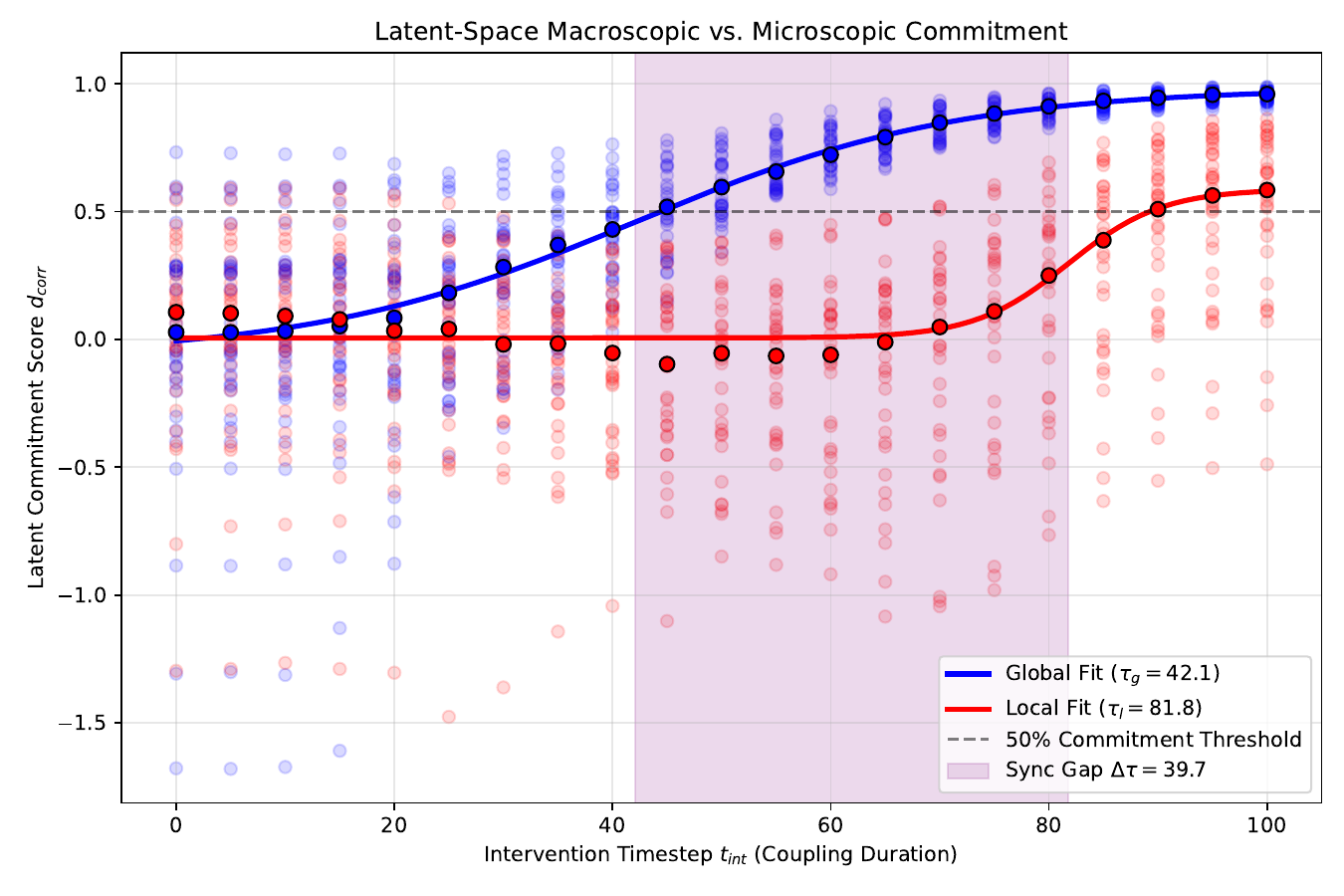}
    \caption{$g=0.9$}
    \end{subfigure}
    \begin{subfigure}[b]{0.8\textwidth}\includegraphics[width=0.99\linewidth]{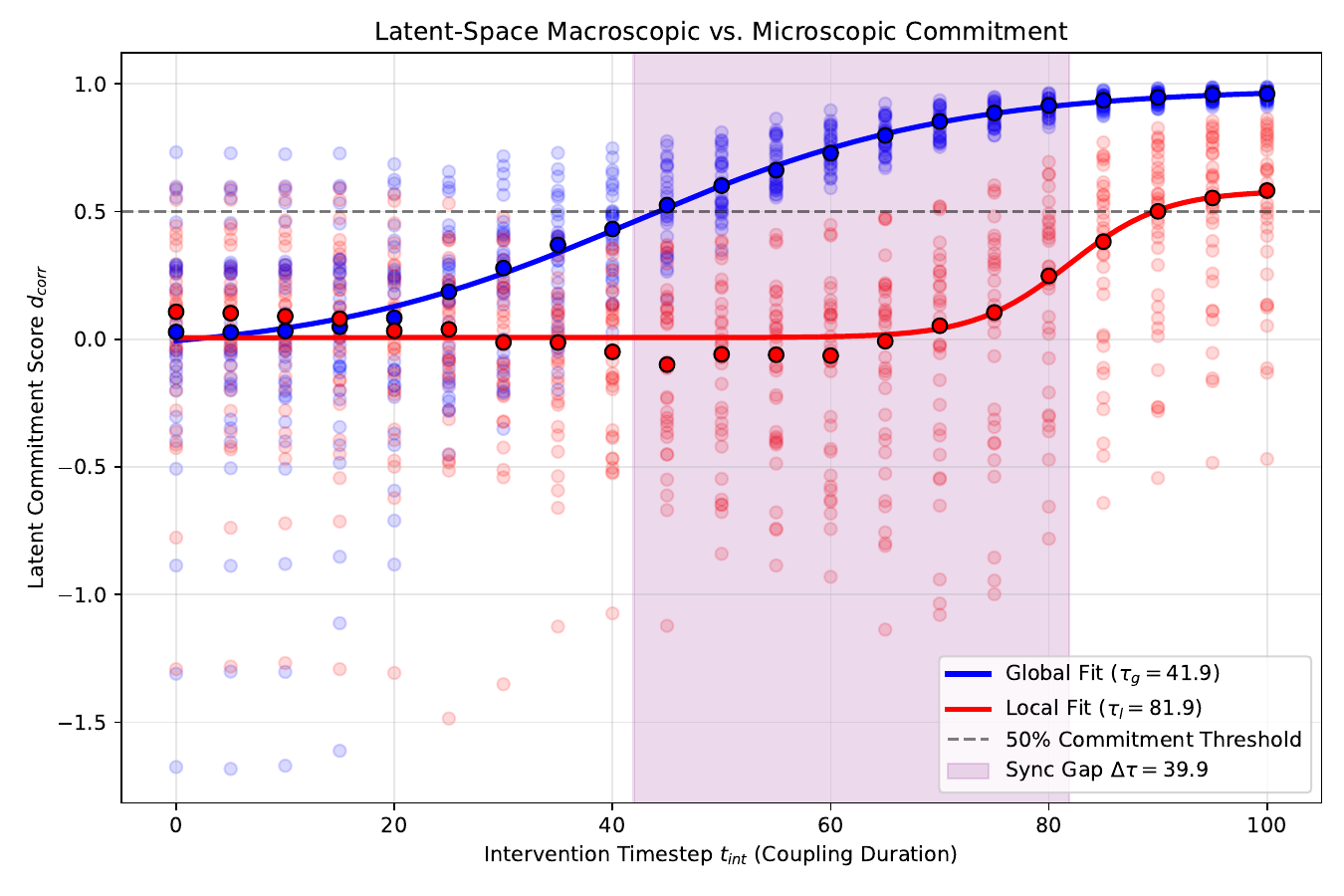}
    \caption{$g=1.0$}
    \label{fig:ComitG09}
    \end{subfigure}
    \caption{Each plot shows a global \eqref{eqn:LowComit} and local \eqref{eqn:HighComit} commitment scores vs coupling strength $g$. These show that coarse output structure stabilizes substantially earlier than fine detail throughout the explored coupling range.}
\end{figure}

\subsection{Protocol II Results}
We now discuss the central experimental result of this paper, given by Protocol II setup \secref{sec:EmpProtII}, which provide the primary mechanistic test of the theory. For a particular coupling strength $g$, we evaluate the normalized fixed basis energies of leading and trailing internal difference modes \eqref{eqn:ProtIILead}--\eqref{eqn:ProtIITrail} at the speciation time identified in Protocol I experiment and sweep the capture layer across Transformer depth. In the case of DiT-XL/2, we sweep through all 28 layers. We use 100 reverse sampling steps with deterministic DDIM sampler $\eta=0$ and we normalize replica attention as in \eqref{eqn:AttnBlocks}, we plot mode mean and $\pm\sigma$ spread from paired seeds. The central qualitative result is a strong coupling dependent suppression of the internal hidden state synchronization gap. 

In the case $g=0$, \figref{fig:SweepG0}, the leading and trailing mode energies remain close
through most of the network but separate sharply in the final blocks, where the trailing
band retains substantially more energy than the leading band. Thus, the internal synchronization gap is strongly depth localized, it is weak in early and middle layers and becomes most evident only near the deepest blocks. The weak coupling case $g=0.1$ exhibits the same qualitative pattern \figref{fig:SweepG01}, although the late layer split is somewhat reduced. Both $g=0$ and $g=0.1$ exhibit an inversion period, roughly between 16th and 22th layer where the local mode is overpassed by the global mode. Especially, for the case of $g=0.0$ these period is characterized by large spread which indicates high dependence on random seeds. We interpret this behavior in the following way, because coupling $g$ is zero or small some replica trajectories diverge wildly causing the energy difference spike. However, later layers damp the chaotic internal variance and reestablish the proper leading/trailing hierarchy 

By contrast, at $g=0.3$ the two curves are already nearly indistinguishable across essentially the full depth of the network \figref{fig:SweepG03}, indicating that moderate coupling is sufficient to suppress most of the internal hierarchy predicted by the spatial routing dominant theory. At $g=0.9$, \figref{fig:SweepG09}, the collapse is even more complete, the leading and trailing mode energies are nearly superposed across depth, with only negligible residual differences.

These depth sweeps support the main theoretical claim of the paper. The effective spatial routing term in the linearized difference channel propagator is weighted by $\rho(g)=(1-g)/(1+g)$, \eqref{eqn:delta-attn-compact}, so the hierarchy between leading and trailing modes should weaken as $g$ increases and should collapse in the strongly coupled regime. The data are consistent with this prediction and further sharpen it. Empirically, the collapse is not confined to the asymptotic regime $g\to 1$, but is already largely realized by moderate coupling.

A second important observation is that the internal gap is concentrated in late blocks. This depth localization is also consistent with the theoretical picture in which layer dependent modulation and residual composition cause the effective difference channel Jacobian to vary strongly across depth. In particular, the results suggest that the deepest transformer blocks are where the leading/trailing hierarchy becomes relevant in the uncoupled or weakly coupled regime, while strong coupling suppresses that hierarchy before it can fully emerge.

\begin{figure}[ht]
    \centering
    \begin{subfigure}[b]{0.99\textwidth}
    \includegraphics[width=0.99\linewidth]{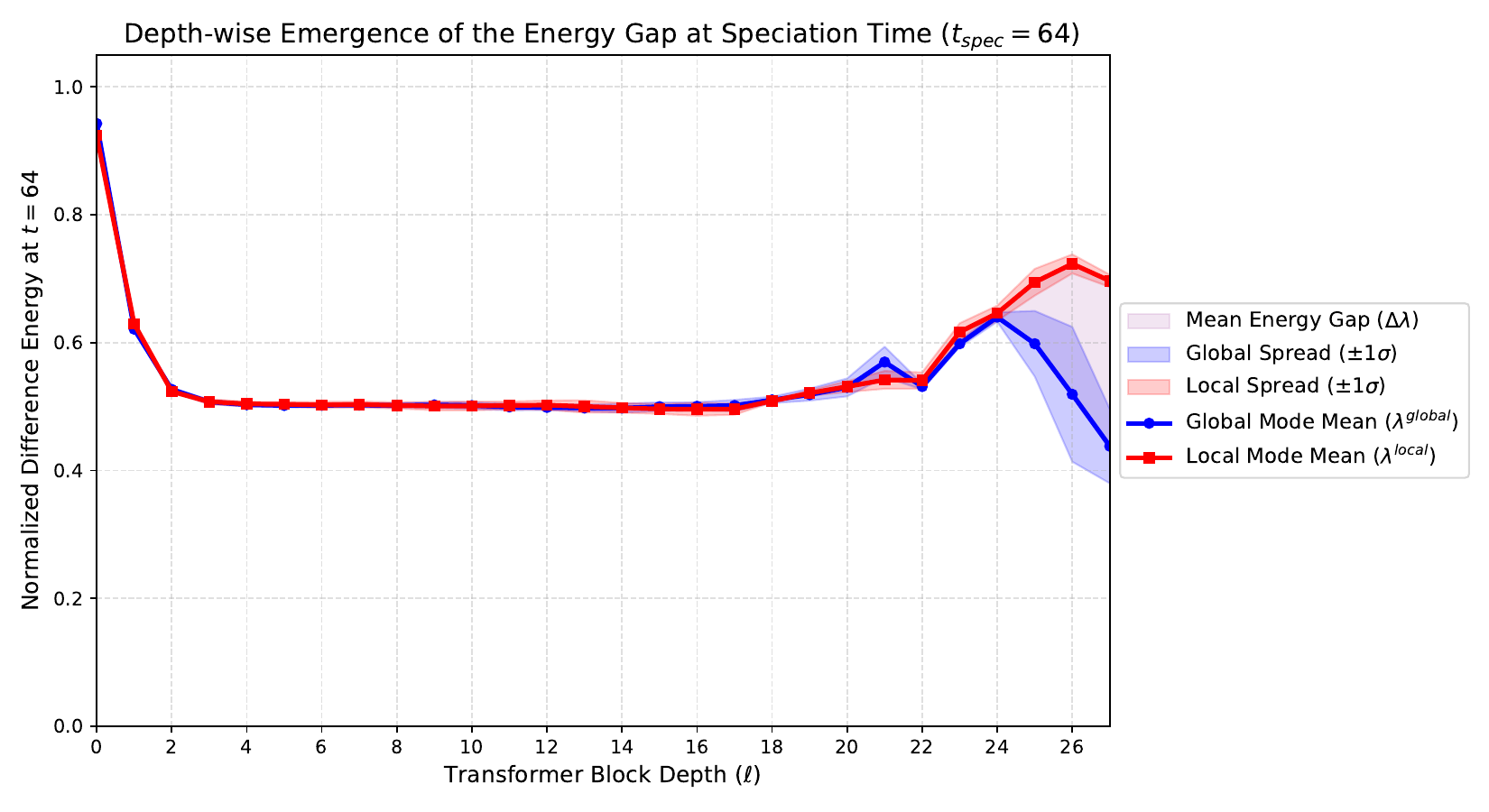}
    \caption{$g=0.1,$ $\tau_{\text{spec}}=64$}
    \label{fig:SweepG01}      
    \end{subfigure}
    \caption{For particular coupling strength $g$ and speciation time we
    sweep across all Transformer layers  to evaluate the normalized fixed basis energies of leading and trailing internal difference
    modes \eqref{eqn:ProtIILead}--\eqref{eqn:ProtIITrail}. These sweeps reveal that the gap is located at deep layers of the Transformer and as coupling strength $g$ increases the gap collapses.}
\end{figure}
\begin{figure}[ht]
    \ContinuedFloat
    \centering
    \begin{subfigure}[b]{0.99\textwidth}
    \includegraphics[width=0.99\linewidth]{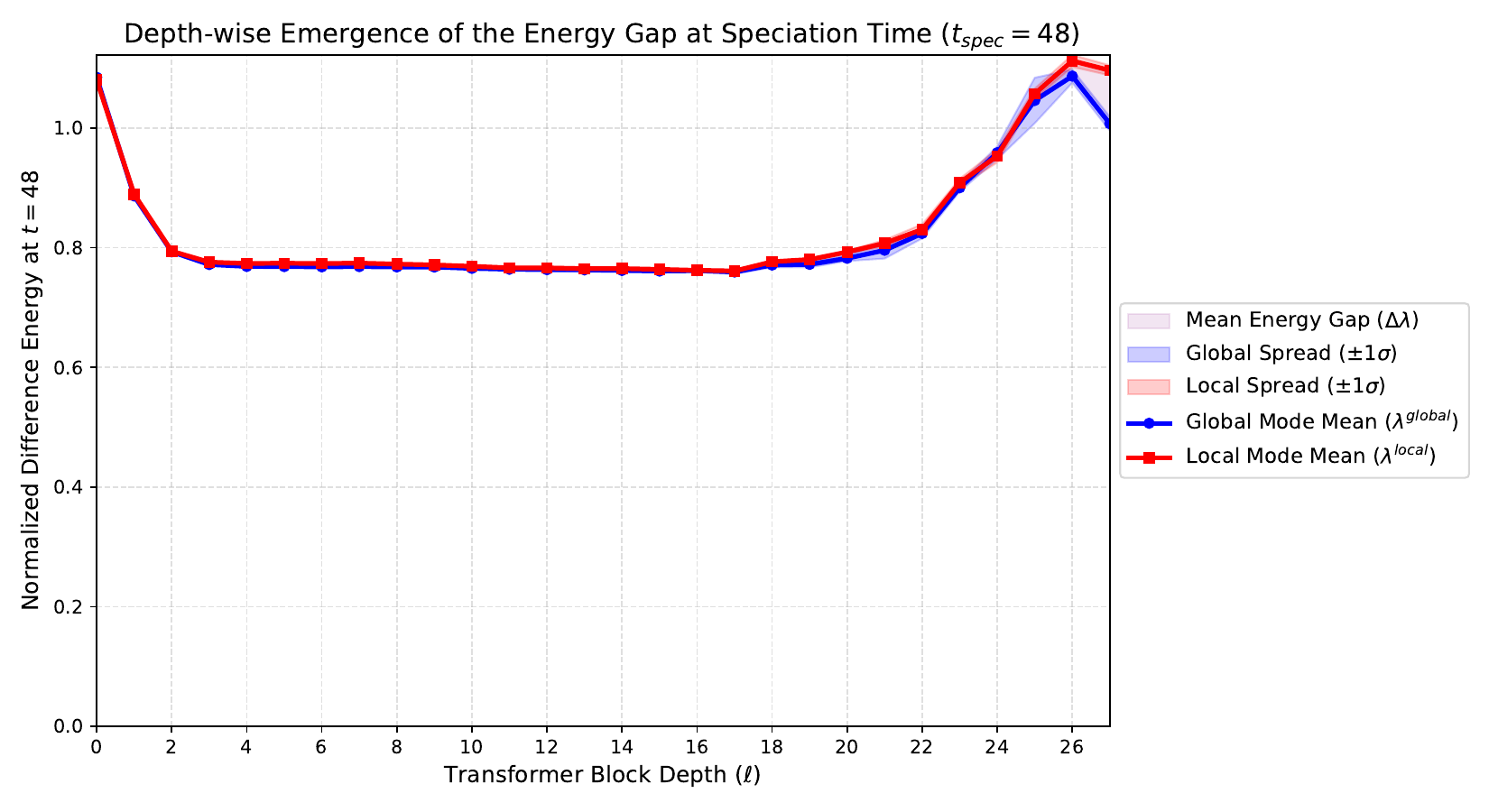}
    \caption{$g=0.3,$ $\tau_{\text{spec}}=48$}
    \label{fig:SweepG03}
    \end{subfigure}
    \begin{subfigure}[b]{0.99\textwidth}
    \includegraphics[width=0.99\linewidth]{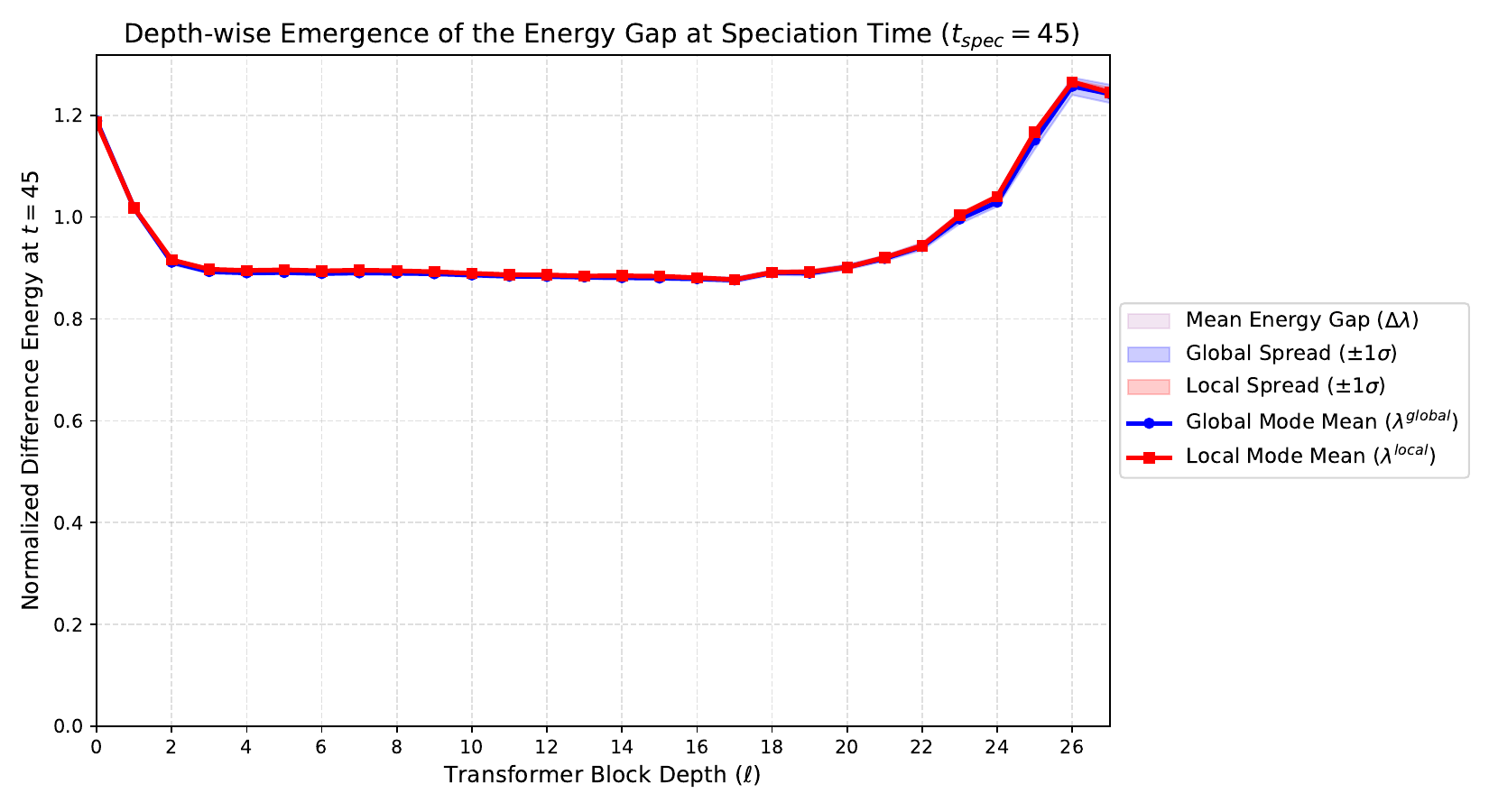}
    \caption{$g=0.9,$ $\tau_{\text{spec}}=45$}
    \label{fig:SweepG09}
    \end{subfigure}
    \caption{For particular coupling strength $g$ and speciation time, we sweep across all Transformer layers  to evaluate the normalized fixed basis energies of leading and trailing internal difference modes \eqref{eqn:ProtIILead}--\eqref{eqn:ProtIITrail}. These sweeps reveal that the gap is located at deep layers of the Transformer and as coupling strength $g$ increases the gap collapses.}
\end{figure}

Taken together, the results support the following picture. Strong symmetric replica coupling sharply suppresses the internal, depth localized hierarchy between leading and trailing difference modes, in agreement with the spatial routing dominant theory. Nevertheless, after coupling is removed, fine output detail still requires a longer period of synchronized evolution than coarse structure in order to survive later independent stochastic refinement. The main novel message is therefore a two level synchronization phenomenon, coupling collapses the internal hidden state hierarchy, while a residual coarse to fine output commitment lag can persist in decoded image space.

\section{Conclusions}
\label{sec:Conclusions}

We have presented a theoretical and empirical study of the synchronization gap in Diffusion Transformers. The main contribution is a controlled framework for analyzing how pretrained DiTs resolve generative ambiguity across spatial scales, connecting the statistical physics picture of coupled diffusion processes to the Transformer architecture.

On the theoretical side, we constructed an explicit architectural realization of symmetric replica coupling inside DiT self-attention \eqref{eqn:AttnBlocks} and derived the linearized attention difference \eqref{eqn:AttnDelta}, which decomposes the network's first order response to the replica difference mode into a spatial routing channel and a pattern modulation channel. We proved, using exact identities of the row stochastic softmax structure \appref{app:RoutingDom}, that for low frequency difference modes the spatial routing term is dominant, the pattern modulation channel exactly cancels the spatially constant component of the common mode value field, while spatial routing preserves it. To obtain a speciation criterion, we modeled the local difference mode distribution as a symmetric two component Gaussian mixture and projected the resulting fixed point equation onto empirical eigenmodes, yielding the scalar self consistency condition \eqref{eqn:speciation-criterion} with a modewise speciation parameter that decomposes into an attention gated SNR \eqref{eqn:snr-def}. Under the routing dominance assumption, the SNR difference between leading and trailing modes scales as $\mathcal{O}(\tfrac{1-g}{1+g})$, predicting the collapse of the internal synchronization gap at strong coupling.

On the empirical side, we tested these predictions on a pretrained DiT-XL/2 model using two complementary protocols. Protocol I \secref{sec:EmpProtI}, based on a decoupling intervention with stochastic sampling, established that the speciation time decreases monotonically with coupling strength \figref{fig:SpecG01}--\figref{fig:specG09} and \tabref{tab:SpecT}. The scale dependent output probe confirmed that global image features commits substantially earlier than local details across all tested coupling strengths, with an output space synchronization gap of $\Delta\tau \approx 39$--$41$ steps that is robust throughout the medium and strong coupling regime \figref{fig:ComitG0}--\figref{fig:ComitG09}. Protocol II \secref{sec:EmpProtII}, based on a sweep across Transformer layers and measurement of internal difference mode energies at the speciation time, revealed three findings. First, the internal synchronization gap exists even at $g = 0$ and is concentrated in the final approximately five Transformer blocks \figref{fig:SweepG0}--\figref{fig:SweepG01}. Second, moderate coupling ($g = 0.3$) is already sufficient to suppress most of the internal leading/trailing hierarchy \figref{fig:SweepG03}. Third, near strong coupling ($g = 0.9$), the leading and trailing mode energies are nearly superposed across the full network depth \figref{fig:SweepG09}.

Taken together, the results reveal a two level synchronization phenomenon. At the internal representation level, coupling collapses the depth localized hierarchy between leading and trailing difference channel modes, consistent with the suppression by the $\tfrac{1-g}{1+g}$ prefactor. At the output level, however, a residual global to local commitment lag persists in decoded image space even under strong coupling, indicating that the VAE decoder and the cumulative effect of many reverse steps introduce additional scale dependent processing that is not captured by the single block linearized theory.

Several limitations of the present work should be noted. First, the local score gain $\gamma_{s,\ell}$, which controls the overall amplitude of the score contribution at each block \eqref{eqn:one-block-map}, was not measured directly from network activations. While the qualitative predictions, mode ordering, gap existence, and gap collapse, depend only on the monotonicity and sign structure of the SNR formula \eqref{eqn:snr-def} and are therefore robust to the precise value of $\gamma_{s,\ell}$, a direct measurement would enable quantitative comparison between predicted and observed speciation times. Second, the mean field projection onto individual empirical modes neglects cross mode couplings that enter at $\mathcal{O}(\|v\|^3)$. Near the bifurcation point where $|u_k|$ is small these corrections are subleading, but away from criticality they may become relevant. Third, the diagonality assumption on $C_{s,\ell}$ \eqref{eqn:modal-projections} in the empirical basis is a modeling approximation whose validity depends on the alignment between the mixture covariance and the initial difference covariance. Fourth, all experiments were conducted on a single architecture (DiT-XL/2) verification across other DiT variants and conditioning modalities would strengthen the generality of the conclusions.
 
The framework developed here opens several directions for future investigation. The observation that the synchronization gap is generated predominantly in the terminal Transformer blocks suggests that targeted interventions at specific layers and reverse time steps could be used to selectively modify the commitment structure of the generative process, with potential applications to controlled generation and concept editing. More broadly, the layerwise propagator $K_g$ \eqref{eqn:KgDef} and the modewise SNR formula \eqref{eqn:snr-def} provide a quantitative language for analyzing how spatial information is routed through the Transformer depth during generation, connecting the macroscopic phenomenology of phase transitions in diffusion models to the microscopic mechanics of self attention. Finally, the explicit connection between the attention gating prefactor and the thermodynamic cost of synchronization, suggested by the dissipative nature of the coupling term, points toward a stochastic thermodynamic  \cite{limmer2024statistical} characterization of the generative process that we leave for future work.

Furthermore, our mechanistic characterization of the synchronization gap suggests a principled interpretation of recent training free acceleration methods based on temporal feature forecasting and feature reuse \cite{liu2025reusing, han2026adaptive}. In particular, our results indicate that commitment of trailing modes is delayed relative to leading modes and that the corresponding internal hierarchy is concentrated in the deepest Transformer blocks. This provides a structural explanation for why temporal approximation can preserve global semantics while degrading local detail, errors introduced during the late stage evolution of trailing modes, especially in terminal blocks, are expected to have a disproportionate effect on fine image fidelity. From this perspective, feature caching strategies should be stage and depth aware, with more reuse in early regimes and exact evaluation retained during late stage refinement and in the final blocks where fine detail commitment is resolved.

\section*{Acknowledgments}
We would like to thank Ori Ganor, David T. Limmer, Wojciech Musial, and Edward Yam for helpful discussions, inspirations and useful comments. We have used Gemini 3.1 Pro for assisted code writing during the development of this paper.  We would like to thank Edward Yam from Vitarai for giving access to an NVIDIA DGX Spark which was used for experiments. This work has been supported by the Leinweber Institute for Theoretical Physics at UC Berkeley.

\begin{appendix}
\section{Detailed Derivation of \eqref{eqn:AttnDelta}}
\label{app:DerAttnDelta}
In this appendix, we suppress the layer index $\ell$. The attention output for replica $A$ is:
\begin{align}
\mathrm{Attn}_g^A = \frac{1}{1+g} \bigl[A_{AA}\,V_A + g\,A_{AB}\,V_B\bigr]\,.
\end{align}
Substituting the first-order expansions:
\begin{align}
    A_{AA}\,V_A &= (A_0 + \delta A^{(+)})(V_0 + \delta V) = A_0 V_0 + A_0\,\delta V + \delta A^{(+)} V_0 + \mathcal{O}(\|v\|^2)\,, \\
    A_{AB}\,V_B &= (A_0 + \delta A^{(-)})(V_0 - \delta V) = A_0 V_0 - A_0\,\delta V + \delta A^{(-)} V_0 + \mathcal{O}(\|v\|^2)\,.
\end{align}
Summing these terms yields:
\begin{align}
\mathrm{Attn}_g^A = \frac{1}{1+g}\Bigl[ (1+g)\,A_0\,V_0 + (1-g)\,A_0\,\delta V + \bigl(\delta A^{(+)} + g\,\delta A^{(-)}\bigr)\,V_0 \Bigr] + \mathcal{O}(\|v\|^2)\,.
\end{align}
By the exact exchange symmetry of the system, $\mathrm{Attn}_g^B$ is obtained by mapping $v \to -v$, which flips the signs of the linear perturbations ($\delta V \to -\delta V$ and $\delta A^{(\pm)} \to -\delta A^{(\pm)}$). Taking the difference $\mathrm{Attn}_g^A - \mathrm{Attn}_g^B$ cancels the base term $A_0 V_0$ and doubles the linear terms, yielding the desired result \eqref{eqn:AttnDelta}.

\section{Routing Dominance for Low Frequency Difference Modes}
\label{app:RoutingDom}

Below we provide a justification for one of a key structural assumptions that we use to analyze the synchronization gap collapse mechanism. In \secref{sec:gap-collapse} we stated that for low frequency difference channel $v$ modes the spatial routing term in the attention difference \eqref{eqn:AttnDelta} dominates the pattern modulation term. The argument has two layers. First, we prove exact identities that follow solely from row wise softmax, and the fact that its entries are nonnegative and normalized to unity. Second, we derive a quantitative residual bound in terms of the effective attention width. As in the theoretical framework section we view token space matrices such as $V_0$ and $\delta V$ as elements of $\R^{N\times d_{\text{model}}}$. 

Let $\mathbf{1}\in\R^N$ denote the token vector of all ones and define the orthogonal projectors
\begin{align}
\label{eq:projectors-routing}
    P_0 := \frac{1}{N}\,\mathbf{1}\mathbf{1}^\top,
    \qquad
    P_\perp := I - P_0.
\end{align}
Thus $P_0$ projects onto the subspace of constant tokens and $P_\perp$ onto the mean free tokens subspace. For any token space matrix $X\in\R^{N\times d_{\mathrm{model}}}$, we have the decomposition
\begin{align}
\label{eq:token-decomp}
    X = P_0 X + P_\perp X.
\end{align}

As defined in \eqref{eqn:AttenMatrix} $A_0\in\R^{N\times N}$ is a row stochastic attention matrix produced by row wise softmax, and $\delta A^{(\pm)}$ is the first order perturbation \eqref{eqn:LogitPert1}-\eqref{eqn:LogitPert2} induced by a logit perturbation $\delta S$. Since blockwise softmax normalizes each of them separately, let us denote both as $\delta A$. Then following is true, for each row $i$, the softmax Jacobian is
\begin{align}
    \frac{\partial [A]_{0,ij}}{\partial [S]_{ik}}
    =
    [A]_{0,ij}\bigl(\delta_{jk}-[A]_{0,ik}\bigr),
\end{align}
where comparing to the notation of \secref{sec:AttnGating} we suppressed channel indices for clarity. To the first order, the perturbation is
\begin{align}
\label{eq:deltaA-row}
    \delta [A]_{ij}
    =
    \sum_{k=1}^{N}
    \frac{\partial [A]_{0,ij}}{\partial [S]_{ik}}\,
    \delta [S]_{ik}
    =
    [A]_{0,ij}
    \left(
        \delta [S]_{ij}
        -
        \sum_{k=1}^{N} [A]_{0,ik}\delta [S]_{ik}
    \right).
\end{align}
Summing over $j$ gives
\begin{align}
    \sum_{j=1}^{N}\delta [A]_{ij}
    &=
    \sum_{j=1}^{N} [A]_{0,ij}\delta [S]_{ij}
    -
    \left(\sum_{k=1}^{N} [A]_{0,ik}\delta [S]_{ik}\right)
    \sum_{j=1}^{N}[A]_{0,ij}
    =0.
\end{align}
Hence $\delta A\,\mathbf{1}=0$ and therefore $\delta A=\delta A P_\perp$ along with normalization fact which implies $A_0P_0=P_0$.

Using these results for $\delta A$ we find
\begin{align}
\label{eqn:PatModSub}
    \delta A V_0=
    \delta A P_\perp V_0,
\end{align}
This result is exact and it shows that pattern modulation can only respond to spatial variation in the common mode value field. It cannot see the constant token component at all.

Now for the spatial routing term we obtain
\begin{align}
    A_0\delta V=P_0\delta V+A_0P_{\perp}\delta V,
\end{align}
so this term preserves the constant token subspace, which was removed by pattern modulation term. 

The structural identities above already imply a strong separation between routing and pattern modulation for coherent token space signals. However, we can provide a stricter bound on pattern modulation residuals. For row $i$ of $A_0$ define effective attention index as 
\begin{align}
\label{eq:neff}
    N_{\mathrm{eff}}^{(i)}
    :=
    \frac{1}{\sum_{j=1}^{N} A_{0,ij}^2},
\end{align}
bounded by $N$, equal to $N_{\mathrm{eff}}^{(i)}=1$ for a single token attention and $N_{\mathrm{eff}}^{(i)}=N$ for a uniform row. Let $\epsilon_j$ denote the rows of the mean free residual $P_\perp V_0$. Then we can write explicitly
\begin{align}
    (\delta A V_0)_i
    =
    \sum_{j=1}^{N}\delta A_{ij}\,\epsilon_j.
\end{align}
Applying Cauchy--Schwarz,
\begin{align}
\label{eq:CS-step}
    \left\|
        \sum_{j=1}^{N}\delta A_{ij}\epsilon_j
    \right\|^2
    \le
    \left(\sum_{j=1}^{N}\delta A_{ij}^2\right)
    \left(\sum_{j=1}^{N}\|\epsilon_j\|^2\right).
\end{align}
From \eqref{eq:deltaA-row},
\begin{align}
    |\delta A_{ij}|
    &=
    A_{0,ij}
    \left|
        \delta S_{ij}
        -
        \sum_{k=1}^{N}A_{0,ik}\delta S_{ik}
    \right|
    \le
    2A_{0,ij}\|\delta S_i\|_\infty,
\end{align}
where
\begin{align}
    \|\delta S_i\|_\infty := \max_{1\le j\le N} |\delta S_{ij}|.
\end{align}
Therefore,
\begin{align}
    \sum_{j=1}^{N}\delta A_{ij}^2
    \le
    4\|\delta S_i\|_\infty^2
    \sum_{j=1}^{N}A_{0,ij}^2
    =
    \frac{4\|\delta S_i\|_\infty^2}{N_{\mathrm{eff}}^{(i)}}.
\end{align}
and that gives per row $i$ bound
\begin{align}
\label{eq:pattern-bound-row}
    \left\|
        (\delta A V_0)_i
    \right\|^2
    \le
    \frac{4\|\delta S_i\|_\infty^2}{N_{\mathrm{eff}}^{(i)}}
    \sum_{j=1}^{N}\|\epsilon_j\|^2.
\end{align}
If we sum over all rows we get a norm bound
\begin{align}
    \|\delta A V_0\| \le  \frac{2\sqrt{N}}{\sqrt{N_{\mathrm{eff}}^{\min}}} \|\delta S\|_{\max} \|P_\perp V_0\|,
\end{align}
where the narrowest attention width is $N_{\mathrm{eff}}^{\min} = \min_{i} N_{\mathrm{eff}}^{(i)}$. Assuming that the common mode field is dominated by the constant token component
\begin{align}
\label{eqn:pattern-bound-final}
    \|\delta A V_0\| \le  \frac{2\sqrt{N}}{\sqrt{N_{\mathrm{eff}}^{\min}}} \|\delta S\|_{\max} \varepsilon_{V_0}\|P_0 V_0\|,
\end{align}
for some constant $0\leq \varepsilon_{V_0}\leq 1$.

For the spatial routing term, the triangle inequality gives
\begin{align}
\label{eq:routing-lower-bound-start}
    \|A_0\delta V\|
    =
    \|P_0\delta V + A_0P_\perp\delta V\|
    \ge
    \|P_0\delta V\| - \|A_0P_\perp\delta V\|.
\end{align}
Assuming that for low frequency a constant component is dominating we can write
\begin{align}
\label{eq:coherence-assumption}
    \|P_\perp\delta V\|
    \le
    \varepsilon_{\delta V}\,\|P_0\delta V\|,
\end{align}
for some constant $0\le \varepsilon_{\delta V}<1$, we refer to this condition as spatial coherency of tokens. Since the norm of $P_\perp$ is one then 
\begin{align}
\|A_0 P_\perp\| \le \lambda_\perp,
\end{align}
where a constant $\lambda_\perp\leq||A_0||$ and altogether for \eqref{eq:routing-lower-bound-start} we have
\begin{align}
\label{eq:routing-lower-bound-final}
    \|A_0\delta V\|
    \ge
    \bigl(1-\lambda_\perp \varepsilon_{\delta V}\bigr)\,
    \|P_0\delta V\|.
\end{align}

Finally, we can combine \eqref{eqn:pattern-bound-final} with \eqref{eq:routing-lower-bound-final} to obtain
\begin{align}
    \label{eq:routing-dominance-main}
    \frac{\|\delta A V_0\|}{\|A_0\delta V\|}
    \le
    \frac{2\sqrt{N}\|\delta S\|_{\max}\,\varepsilon_{V_0}}{\sqrt{N^{\text{min}}_{\text{eff}}}(1-\lambda_\perp \varepsilon_{\delta V})} \frac{\|P_0V_0\|}{\|P_0\delta V\|}.
\end{align}
This equation identifies the regime in which spatial routing dominates pattern modulation in the attention difference. In particular, if
(i) the attention rows are sufficiently diffuse so that $N_{\mathrm{eff}}^{\min}$ is large,
(ii) the common-mode value field $V_0$ is dominated by its constant / slowly varying token component, so that $\varepsilon_{V_0}\ll 1$,
(iii) the difference mode $v$ perturbation $\delta V$ is likewise coherent in token space, so that $\varepsilon_{\delta V}\ll 1$, and
(iv) the attention operator is not too expansive on the mean free token subspace, so that $1-\lambda_\perp \varepsilon_{\delta V}$ remains bounded away from zero,
then the ratio of pattern modulation to spatial routing is small
\begin{align}
    \frac{\|\delta A V_0\|}{\|A_0\delta V\|}
    \ll 1.
\end{align}
Moreover, when $N_{\mathrm{eff}}^{\min}\approx N$, the $N_{\mathrm{eff}}^{-1/2}$ factor provides additional suppression of the pattern-modulation residual. In the DiT-XL/2 setting used in our experiments, $N=1024$, so diffuse attention heads can provide a substantial numerical suppression through the factor $N_{\mathrm{eff}}^{-1/2}$. Thus, in the coherent low frequency regime relevant to early branch formation, the dominant coupling sensitive contribution comes from the spatial routing term rather than the pattern modulation term.

\end{appendix}

\bibliographystyle{JHEP}
\bibliography{diff_ref_url.bib}

\end{document}